\documentclass{article}
\usepackage{graphicx} %
\usepackage[utf8]{inputenc}
\usepackage{amssymb}
\usepackage[margin=1.2in]{geometry}  %

\usepackage{amsmath}
\usepackage{amsthm}

\newtheorem{lemma}{Lemma}
\newtheorem{theorem}{Theorem}

\usepackage{graphicx}
\graphicspath{ {./images/} }
\usepackage{caption}
\usepackage{subcaption}

\usepackage{algorithm}
\usepackage{algpseudocode}

\usepackage{float}

\usepackage{bbm}

\usepackage{enumerate}

\usepackage{dsfont}

\usepackage[skip=10pt plus1pt, indent=20pt]{parskip}

\usepackage{natbib}
\setcitestyle{authoryear,open={(},close={)}}
\usepackage{hyperref}
\hypersetup{
colorlinks   = true,    %
urlcolor     = blue,    %
linkcolor    = blue,    %
citecolor    = blue      %
}
\usepackage{makeidx} %

\title{Fixed-Budget Change Point Identification\\ in Piecewise Constant Bandits}
\author{Joseph Lazzaro \quad\quad Ciara Pike-Burke\\
Department of Mathematics, Imperial College London}
\date{}

\begin{document}

\maketitle
\begin{abstract}
We study the \emph{piecewise constant bandit} problem where the expected reward is a piecewise constant function with one change point (discontinuity) across the action space $[0,1]$ and the learner's aim is to locate the change point. Under the assumption of a fixed exploration budget, we provide the first non-asymptotic analysis of policies designed to locate abrupt changes in the mean reward function under bandit feedback. We study the problem under a large and small budget regime, and for both settings establish lower bounds on the error probability and provide algorithms with near matching upper bounds. Interestingly, our results show a separation in the complexity of the two regimes. 
We then propose a regime adaptive algorithm which is near optimal for both small and large budgets simultaneously.
We complement our theoretical analysis with experimental results in simulated environments to support our findings.
\end{abstract}

\section{Introduction}
In many settings, we are interested in sequentially learning to detect a change point/discontinuity in a piecewise constant function. For example, \citet{ park2021sequentialAdaptiveDesignForJumpRegression,ActiveLearningPiecewiseGP} study the development of materials with physical behaviours which abruptly change under different experimental conditions, such as temperature and pressure. Learning where these changes occur can help predict the quality of production techniques and improve efficiency. However, experiments are expensive and time consuming, so we would like to sequentially choose experimental parameters
to learn where the abrupt changes occur as quickly as possible. The same problem also occurs when mapping out the edge of a cliff on the floor of the ocean \citep{ACPD}. Exploring the whole ocean naively in a grid would be extremely expensive, so we would like to develop strategic ways of minimising the number of times and locations where we measure the depth of the ocean floor.
Developing methods to minimize the number of samples needed to detect a change point
could also help us develop computationally efficient subsampling methods in offline change point analysis, leading to yet more applications in different fields  \citep{fast_offline_cp}.

In this paper, we study this problem, which we refer to as the \emph{Piecewise Constant Bandit Problem}. 
Here, the underlying reward function is a piecewise constant function on $[0,1]$ and the learner's aim is to sequentially select points to query in order to identify the unknown change point $x^*$ as accurately as possible after a fixed number of samples.
We assume that whenever we select a point $x_t \in[0,1]$, we receive a noisy observation of the unknown piecewise constant function at that point.
In contrast to other bandit problems on a continuous action space, in the piecewise constant bandit problem, our goal is to detect where the change in mean occurs rather than identify the optimal arm. Moreover, the abrupt change in the mean reward function violates the smoothness conditions of existing bandit methods for infinite action spaces \citep[e.g.][]{ZoomingAlgo,GPUCB_paper,treeAlgo} meaning that new techniques need to be developed. 
The fixed budget assumption on the number of samples also necessitates the development of new methods. When the budget tends to infinity, some asymptotic methods have been proposed \citep{active_Hall2003,Active_Lan_2009}. However in most practical cases, we only have a finite budget so it is essential to develop a non-asymptotic understanding of the complexity of the problem and develop theoretically justified policies for the fixed budget piecewise constant bandits problem. 

\begin{figure}[t]
\centering
\includegraphics[width=0.5\linewidth]{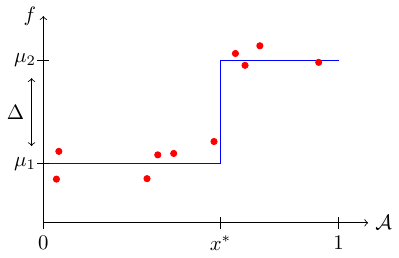}
\vspace{.15in}
\caption{Example of a piecewise constant mean reward function, $f$, across action space $[0,1]$ with change point $x^*$, change in mean of $\Delta$ from $\mu_1$ to $\mu_2$ and 10 arbitrarily chosen noisy observations in red.}
\label{fig_example_mean_fns2}
\vskip -0.2in
\end{figure}

We study the piecewise constant bandit problem in
 environments with exactly one change point across a one dimensional action space and sub-Gaussian noise, see Figure \ref{fig_example_mean_fns2} for an example. While this setting may appear restrictive, it turns out that significant innovation is required to develop optimal methods for this setting, and we hope that these ideas will inspire solutions to more complex problems. We focus on the non-asymptotic fixed budget problem where we are given a fixed number of queries and our aim is to return an estimated change point that minimizes the error probability. 
Finding optimal solutions to the piecewise constant bandit problem is non-trivial. 
Indeed, it requires distributing a finite number of samples across an infinite action space, comparing them to detect a change in mean, and allocating sufficient samples near the unknown change in order to confidently determine its location.
We make the following contributions. 
\textbf{(i)} We characterise the difficulty of the piecewise constant bandits problem for both large \emph{and} small budgets by proving lower bounds that show a separation in difficulty of the two regimes.
This is in contrast to most of the fixed-budget bandits literature, where sufficiently large budgets are explicitly or implicitly assumed \citep[e.g.][]{locatelli16_thresholding,Carpentier2016TightLowerBounds,problem_dependent_threshold}. Our proof techniques are novel and lead to improved lower bounds in related problems such as Thresholding Bandits (see Section \ref{section_largeT_lower_bound}).
\textbf{(ii)} We adapt two elimination algorithms, based on Sequential Halving and Binary Search \citep[e.g.][]{sequential_halving_original,problem_dependent_threshold} to our setting and prove that these have near matching upper bounds for error probability in both regimes.
\textbf{(iii)} We propose a regime adaptive method which is near optimal across both regimes simultaneously.
\textbf{(iv)} We complement our theoretical results with experiments
in simulated environments.

\section{Related Work} \label{section: related work}
\textbf{Pure Exploration in Finite Action Spaces} \quad In fixed budget Best Arm Identification
there is a fixed budget of samples and the aim is to minimize the probability of failing to identify the (approximate) arm with the highest expected reward \citep{MannorLowerBoundPaper,Carpentier2016TightLowerBounds,bubeck_BAI}. This motivates the objective we consider in Section \ref{section:problem setting}.
Other relevant pure exploration problems are Noisy Binary Search \citep{karp_kleinberg}, Binary Classification with Noise \citep{locatelli16_thresholding,Acitve_Castro2008}, and the (Monotonic) Thresholding Bandits problem \citep{problem_dependent_threshold}. In these problems different assumptions are made on the reward function $f$ and noise distribution, however they all aim to locate where $f$ crosses a \emph{known} threshold $\theta$.
This is simpler than the piecewise constant bandit problem since all observations can be compared with the \emph{known} threshold $\theta$. In our case, we do not have a known threshold $\theta$ and so we have to compare observations with each other across an infinite action space to determine where the change in mean occurs, making the problem more challenging. Nonetheless, we are able to extend ideas from \citet{problem_dependent_threshold} to develop nearly minimax optimal algorithms for piecewise constant bandits. 
In clustering with bandit feedback \citep{Yang2022OptimalClustering,yavas2024generalframeworkclusteringdistribution,thuot2024activeclusteringbanditfeedback} arms are sorted in to a known number of clusters, each containing arms with the same expected reward. Unlike our setting, they do not assume any structure in the underlying reward function. Moreover, these clustering problems have not been considered in the fixed budget setting we consider.

\textbf{Infinite Action Spaces}\quad 
Bandit problems with continuous action spaces assume the reward function is linear \citep[e.g.][]{ linear_bandits_yadkori}, convex \citep[e.g.][]{convex_bandits}, Lipschitz \citep[e.g.][]{treeAlgo} 
or smooth enough to be modelled by a Gaussian Process (GP) \citep[e.g.][]{GPUCB_paper}.
In our setting, the mean reward function has an abrupt change, which violates the smoothness assumptions of these  works.

\textbf{Change Points and Non-Stationary Bandits}\quad
Both offline and online change point detection for time series have been well studied in statistics \citep{CP_survey}. 
Online change point analysis has been used in non-stationary bandits. Here, the mean reward of the $K$ arms evolves over time with some abrupt changes
\citep[e.g.][]{Discounted_UCB_non_stationary}.
Conversely, in our setting, the reward changes abruptly over the action space, but is stationary across time.

\textbf{Change Points Across the Action Space} \quad
The active learning literature has studied sequential methods for learning entire piecewise continuous functions \citep[e.g.][]{AdaptiveDesignSupercomputer,park2021sequentialAdaptiveDesignForJumpRegression,ActiveLearningPiecewiseGP}. 
While these papers only provide empirical results, 
\citet{Active_Castro2005} 
develop a two-stage algorithm with  near minimax optimal expected squared $\ell_{2}$ error (in a problem-independent sense and up to $\log$ terms). This measures the accuracy of estimating the entire reward function. In our work, however, we focus on methods for estimating the \emph{locations} of the discontinuities in the reward function. Our PAC methods are  minimax optimal in a \emph{problem-dependent} sense.

 \citet{active_Hall2003} and \citet{Active_Lan_2009} consider fixed budget multi-stage methods to sequentially estimate discontinuities in piecewise smooth functions with exactly one change point. Their first stage uses a portion of the budget to sample evenly across the action space to construct a confidence interval (CI) for the change point. In the following stages this process is repeated within the previous CI (backtracking to an earlier CI if they no longer detect a change). The theoretical guarantees provided only hold as the budget tends to infinity, which clashes with the motivation of fixed budget problems with an extremely limited number of samples. In certain finite-budget problems, their methods perform suboptimally. In particular, if there are not many samples one side of the change point, the initial CI for the change point can be unreliable and the final estimate performs poorly (i.e. the algorithm is unstable for changes near the boundary - see Section \ref{section_experiments}).
 Moreover, their results are worst case and do not provide any insights into how performance changes depending on the problem instance. Thus we provide the first
 non-asymptotic
 problem dependent bounds on the performance of algorithms for  piecewise constant bandits which apply for most realistic budgets. We also provide near matching problem-dependent lower bounds demonstrating that our methods are near optimal, and correctly adapt to the difficulty of the problem.

 \citet{ACPD} consider a class of active change point detection problems, where the aim is to locate changes in reward functions that are piecewise constant, piecewise linear, 
 or contain other types of change points.
 They propose an anytime meta-algorithm, ACPD, which requires a statistical model for the type of change point and noise distribution to calculate ``change scores" for different regions of the space. They run Bayesian Optimisation using these scores at each iteration. 
No theoretical guarantees
for
ACPD have been provided. Our simulations in Section~\ref{section_experiments} demonstrate that, while ACPD can 
perform well on `easy' problems, there are settings where our theoretically grounded methods perform significantly better, 
while also being computationally cheaper than ACPD.

\section{Problem Setting}\label{section:problem setting}
We consider the piecewise constant bandit problem with a fixed budget $T$. Here, in each round $t=1,\dots, T$, we choose an action $x_t \in \mathcal{A}=[0,1]$ and observe a reward $y_t \in \mathbb{R}$. We consider a set of environments 
$V$
where the mean reward function $f:[0,1] \mapsto \mathbb{R}$ is a piecewise constant function with exactly one change point $x^* \in [0,1)$. Then, the reward we observe from playing $x_t$ in round $t$ is
\begin{equation*}
    y_t=f(x_t)+\epsilon_t,
\end{equation*}
where $f(x)=\mu_1\mathbb{I}\{x\leq x^*\}+\mu_2\mathbb{I}\{x > x^*\}$ is a piecewise constant function with $\mu_1\neq\mu_2 \in \mathbb{R}$. We assume the random noise, $\epsilon_t$, is i.i.d and $\sigma^2$ sub-Gaussian with mean-zero.
Importantly, we assume that the values of $\mu_1\neq\mu_2 \in \mathbb{R}$, $\sigma^2 \in [0,\infty)$, and $ x^* \in [0,1)$ are unknown to the learner. 
We subsequently denote the change in mean reward as $\Delta = |\mu_1-\mu_2|$ and $V(\Delta,\sigma)$ as the set of environments with change in mean reward
$\Delta$, 
$\sigma^2$ sub-Gaussian noise, and any change point $x^* \in [0,1)$. 

Our goal is to estimate the change point, $x^*$, which separates the two reward distributions across $[0,1]$. In particular, given some budget $T$, our objective is to generate an estimate after $T$ rounds, $\hat{x}_T$, which is close to the true unknown change point with high probability. Finding a change point exactly is impossible in a continuous action space, so we define an \emph{acceptable tolerance} $\eta>0$ and aim to return an estimate within $\eta$ of the true change point. Let $\Pi$ be the set of all policies which return an estimate for the change point $\hat{x}_T$ after budget $T$. We aim to find a policy $\pi \in \Pi$ such that in environment $v \in V$ with change point $x^*_v$,
\begin{equation}\label{eqn_objective}
    \mathbb{P}_{\pi,v}(|\hat{x}_T-x^*_v|<\eta) > 1-\delta
\end{equation}
with $\delta$ as small as possible. Here $\mathbb{P}_{\pi,v}$ is the measure induced by the interactions between the policy $\pi$ and environment $v$ (we henceforth drop the subscript whenever it is clear which policy and environment we are referring to).
We assume the learner is given  fixed values for $T$ and $\eta$, and their goal is to develop a policy that returns an estimate $\hat{x}_T$ which satisfies \eqref{eqn_objective} with $\delta$ as small as possible. %
Equivalently, we want to minimise the probability that we fail to estimate the change point with sufficient accuracy after a given number of observations. 
Defining the failure event as $F_{T,v,\eta}=\{|\hat{x}_T - x^*_v|\geq\eta\}$, we can state this objective as minimizing 
$\mathbb{P}_{\pi,v}(F_{T,v,\eta})$.
Note that this objective is similar to 
those
seen in other pure exploration problems such as PAC best arm identification \citep[e.g.][]{MannorLowerBoundPaper}. 
This objective is also practically relevant as a practitioner will only care about a particular level of precision across an infinite space, hence it is natural to include a pre-specified tolerance $\eta$.

\section{Large Budget}\label{section_largeT}

\begin{figure*}[t]
\centering
\includegraphics[width=0.9\linewidth]{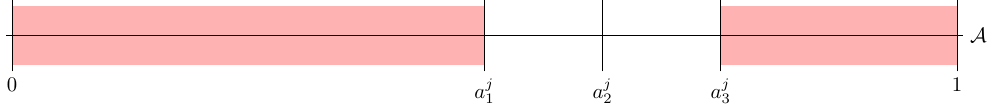}
\includegraphics[width=0.9\linewidth]{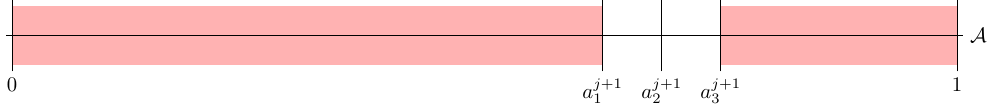}
\vspace{.15in}
\caption{Example illustration of action space $\mathcal{A}=[0,1]$ in (Top) phase $j$ with sampling points $\mathcal{A}^j = \{0,a_1^j,a_2^j,a_3^j,1\}$ and (Bottom) phase $j+1$ with sampling points $\mathcal{A}^{j+1} = \{0,a_1^{j+1},a_2^{j+1},a_3^{j+1},1\}$, where the shaded regions have been eliminated. In this example, $E_{R,j}$ held and region $[a_1^j,a_2^j)$ was eliminated in phase $j$.}
\label{fig_example_phases}
\vskip -0.2in
\end{figure*}

We first study
the setting where the budget, $T$,  is large enough to efficiently explore the space and accurately locate the change point, while still considering non-asymptotic methods. We consider an extension of the binary search with backtracking algorithm which was used for Monotonic Thresholding Bandits and Noisy Binary Search \citep{problem_dependent_threshold,karp_kleinberg}. Informally, the idea is to split our budget up into phases and in each phase sample the leftmost, mid, and rightmost point in the action space repeatedly. If there is more evidence that a change in mean occurs on the left half than the right half, then we eliminate the right half of the space. Similarly, if there is more evidence that the change occurs on the right half, we eliminate the left half of the space.
If, however, at any point we think the remaining region no longer contains the change point, we go back a step and undo our previous elimination (i.e.  backtrack).

Naively applying the binary search with backtracking algorithm from \citet{problem_dependent_threshold} to our setting would not work. 
In Thresholding Bandits, all observation could simply be compared with a known threshold $\theta$. However, in our case we need to compare \emph{unknown} reward distributions across the action space with each other in order to find a change point. 
In particular, if we want to correctly identify when a change point is \emph{not} in some region $B \subset \mathcal{A}$ with high probability, then it is crucial to have data from \emph{both} sides of the change point. 
Therefore, we additionally sample outside of region $B$, at the boundaries of the action space $0$ and $1$, in each phase. 
We call our algorithm \textbf{Sequential Halving with Backtracking (SHB)}. SHB is detailed in Algorithm~\ref{algo:combined} by setting \texttt{backtracking}=True (a standalone version is in Appendix \ref{appendix: algo def}).

\begin{algorithm}[t]
\caption{Sequential halving (with backtracking) -  SH(B)}\label{algo:combined}
\begin{algorithmic}[1]
\State {\bfseries Input:} $\eta \in (0,1/2)$, budget $T$, and 
\State {\bfseries Input:} \texttt{backtracking} = True or False
\State $\mathcal{A}^1 = \{0,a_1^1,a_2^1,a_3^1,1\} \gets \{0,0,1/2,1,1\}$
\If{\texttt{backtracking}}
\State $J \gets \lceil 6\log  (1/2\eta) \rceil,\quad t_j \gets \lfloor \frac{T}{5J}\rfloor$
\Else
\State $J \gets \lceil \log _2 (1/2\eta) \rceil,\quad t_j \gets \lfloor \frac{T}{3J}\rfloor$ 
\EndIf
\For{phase $j$ in $1, ... , J$}
\State Play actions $a_1^j,a_2^j,a_3^j$, \, each $t_j$ times
\If{\texttt{backtracking}}
\State Play actions $0,1$, \, each $t_j$ times
\EndIf
\If{\texttt{backtracking} \textbf{and} $E_{P,j}$ from \eqref{event_E_P} holds}
\State $\mathcal{A}^{j+1} \leftarrow P(\mathcal{A}^{j})$\Comment{Backtrack}
\ElsIf{$E_{R,j}$ from \eqref{event_E_R} holds}
    \State $\mathcal{A}^{j+1} \leftarrow R(\mathcal{A}^{j})$ \Comment{Zoom in to the right}
\ElsIf{$E_{L,j}$ from \eqref{event_E_L} holds}
    \State $\mathcal{A}^{j+1} \leftarrow L(\mathcal{A}^{j})$ \Comment{Zoom in to the left}
\EndIf
\EndFor
\State \textbf{Return:} $\hat{x}_T = a_2^{J+1}$
\end{algorithmic}
\end{algorithm}

In more detail, in SHB, we split our budget $T$ into $J = \lceil 6\log  (1/2\eta) \rceil$ phases. 
In each phase $j \in \{1,..,J\}$, we define the set of sampling points as $\mathcal{A}^j = \{0,a_1^j,a_2^j,a_3^j,1\}$ and begin the phase by playing each action in $\mathcal{A}^j$  $t_j = \lfloor \frac{T}{5J}\rfloor$ times. The set of sampling points in $\mathcal{A}^j$ consist of the endpoints and midpoint of the non-eliminated region, together with the extreme points 0,1 needed for backtracking. This is illustrated in Figure \ref{fig_example_phases}. 
In phase 1 we have not eliminated anything yet, so we initialise with $a_1^j=0,a_2^j=1/2,a_3^j=1$.
In every phase $j=1,\dots, J$, we compute the empirical mean of the observations from playing each of the five actions actions $t_j$ times and denote them $\hat{\mu}_{0,t_j},\hat{\mu}_{a_1^j,t_j},\hat{\mu}_{a_2^j,t_j},\hat{\mu}_{a_3^j,t_j},\hat{\mu}_{1,t_j}$, respectively. 
These estimates help identify which half of the remaining action space $[a^j_1,a^j_3)$ contains the change point. 

Firstly, we determine whether there is more evidence for $x^*$ being in the left or right half of this remaining space in phase $j$, i.e we determine whether $x^*$ is in $[a^j_1,a^j_2)$ or $[a^j_2,a^j_3)$.
In particular if the event%
\begin{align}
    E_{R,j} &= \left\{ |\hat{\mu}_{a_1^j,t_j} - \hat{\mu}_{a_2^j,t_j}| < |\hat{\mu}_{a_2^j,t_j} - \hat{\mu}_{a_3^j,t_j}| \right\}, \label{event_E_R}
\end{align}
holds then we suspect that the change point is in the right half of the remaining action space, namely $x^* \in [a_2^j,a^j_3)$. This is because $E_{R,j}$ occurs when we observe a bigger change in (empirical) mean reward between actions $a_2^j$ and $a^j_3$, than between $a^j_1$ and $a^j_2$, suggesting that the change in distribution occurred between $a_2^j$ and $a^j_3$. (It turns out that this intuition is theoretically justified from change point analysis, see Appendix \ref{appendix_cps}.)  In such a case, we `zoom in to the right' by eliminating the left half of the remaining action space and define our actions for the next phase as %
$\mathcal{A}^{j+1} = R(\mathcal{A}^{j})$, for the operator
\begin{align*}
    R(\{0,a_1^j,a_2^j,a_3^j,1\}) &= \{0,a_2^j,(a_2^j+a_3^j)/2,a_3^j,1\},
\end{align*}
so $\mathcal{A}^{j+1}=\{0,a_1^{j+1},a_2^{j+1},a_3^{j+1},1\}$ for $a_1^{j+1}=a_2^j, a_2^{j+1} = (a_2^j+a_3^j)/2, a_3^{j+1}=a_3^j$.
See Figure \ref{fig_example_phases} for an illustration.

On the other hand, if the converse is true and event 
\begin{align}
    E_{L,j} &= \left\{ |\hat{\mu}_{a_1^j,t_j} - \hat{\mu}_{a_2^j,t_j}| \geq |\hat{\mu}_{a_2^j,t_j} - \hat{\mu}_{a_3^j,t_j}| \right\} \label{event_E_L}
\end{align}
holds, this suggests that the change point is in the left half of the remaining action space, $x^* \in [a^j_1,a^j_2)$.
In this case, we `zoom in to the left' by eliminating the right half of the space, $[a^j_2,a_3^j)$, and update our actions for the next phase as
$\mathcal{A}^{j+1} = L(\mathcal{A}^{j})$ for the operator %
\begin{align*}
    L(\{0,a_1^j,a_2^j,a_3^j,1\}) &= \{0,a_1^j,(a_1^j+a_2^j)/2,a_2^j,1\}.
\end{align*}
It is, of course, possible that the change point is not within our remaining action space $[a^j_1,a^j_3)$ at all. Hence, before we even consider checking events $E_{L,j}$ and $E_{R,j}$,
we consider the event $E_{P,j}$,
\begin{align}
    E_{P,j} &= \left\{Q_1 < \frac{3}{4}\max \left( Q_2 ,\, Q_3     \right) \right\}\label{event_E_P}\\
    \text{ for } \qquad Q_1 &= \left|\frac{\hat{\mu}_{0,t_j}+ \hat{\mu}_{a_1^j,t_j}}{2} - \frac{\hat{\mu}_{a_3^j,t_j}+\hat{\mu}_{1,t_j}}{2}\right|, \nonumber\\
    Q_2 &= \left| \frac{\hat{\mu}_{0,t_j} + \hat{\mu}_{a_1^j,t_j} + \hat{\mu}_{a_3^j,t_j}}{3} - \hat{\mu}_{1,t_j}\right|,\nonumber\\
    Q_3 &= \left| \hat{\mu}_{0,t_j} - \frac{\hat{\mu}_{a_1^j,t_j} + \hat{\mu}_{a_3^j,t_j} + \hat{\mu}_{1,t_j}}{3}\right|. \nonumber
\end{align}
If $E_{P,j}$ holds then we suspect that $x^* \notin [a_1^j,a_3^j)$ so we believe the change point is elsewhere. Intuitively $E_{P,j}$ occurs when the change in empirical mean is smaller across the region $[a^j_1,a^j_3)$ than elsewhere, see Appendix \ref{appendix_cps} for more details.
If $E_{P,j}$ holds then we `zoom out' one step by un-doing the previous elimination and update our actions for the next phase as $\mathcal{A}^{j+1} = P(\mathcal{A}^{j})$. The operator $P$ returns the sampling points $\mathcal{A}^{i}$ from which we previously zoomed in to get to $\mathcal{A}^{j}$. More formally, we define $P(\mathcal{A}^{j})$ as an operator that outputs a set of actions from a previous phase, $\mathcal{A}^{i}$ such that $i \leq j-1$ and we zoomed in from $\mathcal{A}_i$ to obtain our current set of actions $\mathcal{A}^{j}$. Namely $P(\mathcal{A}^{j}) = \mathcal{A}^i$ for an $i \in \{1,\dots, J\}$ such that $\mathcal{A}^{j} = R(\mathcal{A}^{i})$ or $\mathcal{A}^{j} = L(\mathcal{A}^{i})$.

At the end of the $J$ phases, we are left with the set $\mathcal{A}^{J+1} = \{0,a_1^{J+1},a_2^{J+1},a_3^{J+1},1\}$ representing the left-most point, midpoint, and right-most point of the remaining action space that has not been eliminated. We estimate the change point, $x^*$, by taking the midpoint of this remaining region, namely 
$\hat{x}_T = a_2^{J+1}$.

\subsection{Upper Bound}
We upper bound the probability that the SHB algorithm fails to estimate the true change point $x^*$ up to the acceptable tolerance $\eta$. Namely, we upper bound $\mathbb{P}(F_{T,v,\eta})$, with failure event $F_{T,v,\eta}=\{|\hat{x}_T - x^*_v|\geq\eta\}$.

 \begin{theorem}\label{fixed budget upper bound backtracking}
    Let $\eta < 1/4$. Consider SHB in an environment $v \in V(\Delta,\sigma)$. 
    Then, %
    for $T> 60 \log (1/2\eta)$, %
    \begin{equation*}
        \mathbb{P}(F_{T,v,\eta})\leq \exp\left(-\frac{\Delta^2}{600\sigma^2}T + 13\log(1/2\eta)\right).
    \end{equation*}
 \end{theorem} 
 \begin{proof}  We sketch the proof here, leaving details for Appendix \ref{app_largeT_upper_proofs}. Note that, since SHB is an extension of \citep{problem_dependent_threshold}, proofs for some technical lemmas are similar to existing work. In particular, we first note that we don't need to make the correct decision in eliminating/backtracking in every phase. As long as we make correct decisions in at least 3/4 of the phases, then our final estimate for the change point, $\hat{x}_T$, will be within $\eta$ of $x^*$. However, our elimination/backtracking criteria are more involved than \citet{problem_dependent_threshold}. Hence we must develop new techniques to show that we make the correct decision in each phase with high probability, regardless of the position of $x^*$.
 \end{proof}
 
Note that $\log(1/2\eta)$ only appears additively in the exponent in Theorem~\ref{fixed budget upper bound backtracking}. 
This log dependence on $\eta$ has been attained in related settings such as Thresholding Bandits or Active Binary Classification \citep{problem_dependent_threshold, Acitve_Castro2008}. However, unlike these previous works, we show that this log dependence on $\eta$ is unimprovable for larger budgets in Section \ref{section_largeT_lower_bound}, and we will show in Section~\ref{section small budget}  that a different dependence on $\eta$ appears for small budgets. %

\subsection{Lower Bound} \label{section_largeT_lower_bound}
We provide a lower bound to show that the error probability from our SHB algorithm is minimax optimal up to constants, 
for \textbf{large budgets} $T \geq T_1 := \frac{\sigma^2}{\Delta^2}(1.59\log(\lfloor\frac{1}{2\eta}\rfloor)-2\log(2))$.
Note the upper bound in Theorem \ref{fixed budget upper bound backtracking} holds for any sub-Gaussian reward distributions, whereas in this section we assume that the rewards are Gaussian.

\begin{theorem}\label{corr_large_budget_lb}
     Let $\Bar{V} \subset V(\Delta,\sigma)$ be the set of environments with change in mean $\Delta$ and Gaussian random noise with variance $\sigma^2$.
     Then, for $T \geq T_1$,
    \begin{align*}
        \inf_{\pi \in \Pi} \sup_{v \in \Bar{V}} \mathbb{P}_{v,\pi} (F_{T,v,\eta}) \geq & \\
        \frac{1}{8}\exp \bigg[ -\frac{\Delta^2}{2\sigma^2}T + \frac{1}{2} & \log \left( \frac{1}{2} \left( \left\lfloor \frac{1}{2\eta} \right\rfloor -1 \right) \right) \bigg] .
    \end{align*}
\end{theorem}
\begin{proof}
We sketch the proof here, see Appendix \ref{app_largeT_lowerbound_proofs} for full proof.
Suppose we have a policy $\pi$ which has reasonably good failure probability regardless of the environment, namely %
$\forall v \in \Bar{V},\, \mathbb{P}_{v,\pi} (F_{T,v,\eta}) < 1/2$. We would expect this policy to explore the action space sufficiently well. Then, using a sequence of change-of-measure arguments between three carefully chosen environments, we show
such a policy $\pi$ will always make a mistake with some probability (involving $\eta$) in at least one environment.
We then extend this to show that for \emph{any} policy $\pi \in \Pi$, when the budget is sufficiently large we must incur the stated failure 
probability. 
\end{proof}

Comparing the upper bound in Theorem \ref{fixed budget upper bound backtracking} and the minimax lower bound in Theorem \ref{corr_large_budget_lb}, we see that the terms in the exponent of both the upper and lower bounds match up to constants.
Hence for large budgets, $T>T_1$, SHB is minimax optimal up to constants. It is also interesting to note that in both the upper and lower bounds, $\eta$ only has an additive effect in the exponent, and the $\eta$ terms do not scale multiplicatively with $T$. This means that as $\eta$ becomes very small (i.e. when we need our estimates' accuracy to be very high), this only has an additive effect in the exponent of the probability of failure, and does not affect the \textit{rate} at which the failure probabilities decay with increased $T$. 
In the simpler Monotonic Thresholding Bandits Problem \citep{problem_dependent_threshold, Acitve_Castro2008}, a similar additive dependence on $\eta$  ($\eta \approx 1/2K$ %
in their setting) has appeared in upper bounds on the error probability. However, there is currently no lower bound for Thresholding Bandits to show that this dependence on $\eta$ is unavoidable. 
As a consequence of Theorem~\ref{corr_large_budget_lb}, we can provide tighter lower bounds for Thresholding Bandits with large budgets 
which formalises the effect of $\eta$, and may be of independent interest.

\section{Small Budget} \label{section small budget}

For problems where the budget is small and gives little time for exploration, it is natural to consider omitting the exploratory backtracking actions from SHB to provide a more exploitative algorithm. We show that the resulting \textbf{Sequential Halving (SH)} algorithm (Algorithm~\ref{algo:combined} with \texttt{backtracking}=False) is sufficient to attain near-optimality 
for \textbf{small budgets} $T<T_1:=\frac{\sigma^2}{\Delta^2}(1.59\log(\lfloor\frac{1}{2\eta}\rfloor)-2\log(2))$.
The SH algorithm is written explicitly in %
Appendix \ref{appendix: algo def}.

In SH, we split our budget up into  $J=\lceil \log _2 (1/2\eta) \rceil$ phases. Then in every phase, using only samples from the leftmost, mid, and rightmost points of the remaining action space, we eliminate half of this remaining action space. In particular, when event $E_{R,j}$ holds (defined the same as in \eqref{event_E_R}) we eliminate the left half of the action space. When $E_{L,j}$ holds (again, same as \eqref{event_E_L}) we eliminate the right half of the action space. By our choice of the number of phases, the width of the final region will be less than $2\eta$. Hence, we estimate the midpoint of this region to be the change point. 

In this SH algorithm we are more exploitative and avoid additionally sampling the actions $0,1$, as needed for backtracking in SHB, which gives better performance for smaller budgets. However, we will see this comes at the cost of worse error probabilities for larger budget problems. Note that SH requires $T \geq 3\lceil \log _2 (1/2\eta) \rceil$ in order for there to be at least one sample for each action in every phase. This assumption is reasonable since even in the noiseless setting with known $\mu_1,\mu_2$, the minimum number of samples required to guarantee
$|\hat{x}_T-x^*| \leq \eta$\, is $T \geq \lceil \log _2 (1/2\eta) \rceil$ \citep{bisection_optimal}.

\subsection{Upper Bound}
The failure probability %
of the SH algorithm is bounded in Theorem \ref{fixed budget upper bound}, with proofs in Appendix \ref{app_smallT_upperbound_proofs}.  %

\begin{theorem} \label{fixed budget upper bound}
    Under the SH algorithm in an environment $v \in V(\Delta,\sigma)$, for $T \geq 3\lceil \log _2 (1/2\eta) \rceil$, and $F_{T,v,\eta} = \{|\hat{x}_T - x^*_v| \geq \eta\}$, 
    $$\mathbb{P}(F_{T,v,\eta})<2\left\lceil \log _2 \left(\frac{1}{2\eta}\right) \right\rceil \exp\left(\frac{-T\Delta^2}{36\sigma^2 \log_2(1/2\eta)}\right).$$
\end{theorem}

It is important to note that in Theorem \ref{fixed budget upper bound} the $\log_2(1/2\eta)$ term is in the denominator of the exponent, multiplying the leading term. This comes from the fact that, in order to achieve our objective of $|\hat{x}_T-x^*| \leq \eta$ with SH, we need to eliminate the correct half of the action space in \textit{every} phase (unlike in SHB where by backtracking we need only make the correct decision in some proportion of the phases - see Appendix~\ref{app_largeT_upper_proofs}). Therefore for SH, $\eta$ \emph{does affect the rate} at which the failure probability decreases.
This is in contrast to Theorem \ref{fixed budget upper bound backtracking} and Theorem \ref{corr_large_budget_lb} where $\eta$'s involvement is only additive in the exponent and does not affect the rate. This indicates SH is suboptimal for large budgets. 
However, we now show that 
SH is minimax optimal up to constants in the small budget regime. %

\subsection{Lower Bound}
To understand the influence of $\eta$ on the difficulty of the small budget problem, we consider covering/Fano arguments similar to Chapter 15 of \citet{wainwright_2019}. From this we obtain a minimax lower bound
 shown in Theorem \ref{lower_bound_exp_form}, the proof of which is in Appendix \ref{app_smallT_lowerbound_proofs}. By comparing the upper bound for SH in Theorem \ref{fixed budget upper bound} with the lower bound in Theorem \ref{lower_bound_exp_form}, we see that SH is near minimax-optimal in the small budget regime, 
$T<T_1$, 
up to an additive $\log \log(1/\eta)$ term in the exponent.
\begin{theorem}\label{lower_bound_exp_form}
Let $\Bar{V} \subset V(\Delta,\sigma)$ be the set of environments with change in mean $\Delta$ and Gaussian random noise with variance $\sigma^2$. 
    When $T<T_1$,
    we have
    \begin{equation*}
        \inf_{\pi \in \Pi} \sup_{v \in \Bar{V}} \, \mathbb{P}_{\pi,v}(F_{T,v,\eta}) \geq \exp\left(-\frac{\Delta^2 T+ 2\sigma^2\log(2)}{ \sigma^2\log(\lfloor\frac{1}{2\eta}\rfloor)}\right).
    \end{equation*}
\end{theorem}

Theorem~\ref{lower_bound_exp_form} holds for hold for small budgets,  $T<T_1:=\frac{\sigma^2}{\Delta^2}(1.59\log(\lfloor1/2\eta\rfloor)-2\log(2))$ while Theorem~\ref{corr_large_budget_lb} holds for large budgets $T \geq T_1$. These results together give a full characterisation of the difficulty of the fixed budget piecewise constant bandit problem. Interestingly they show a separation in the achievable error probability (with respect to $\eta$) depending on the  budget regime.
Similar separations in complexity have been observed in active learning \citep{discontinuity_paper_active_learning}, although those results hold for fixed confidence problems with a labelling oracle, 
which we do not have in our setting.

\section{Adaptive Algorithm} \label{section_regime_identification}
Dependent on what regime we are in, it may be better to use SH or SHB.
For example, our theoretical results show that 
when T is smaller than a threshold of order $\frac{\sigma^2}{\Delta^2} \log(1/\eta)$ our guarantee for SH (Theorem \ref{fixed budget upper bound}) is better than SHB (Theorem \ref{fixed budget upper bound backtracking}), suggesting SH is better suited for such smaller budgets.
The converse can hold when we consider larger budgets. Note that while these observations come from comparing upper bounds on the performance of SH and SHB, we see experimentally that these conclusions hold (Section \ref{section_experiments}). Moreover, our lower bounds  show that our guarantees are tight up to constants or $\log\log(1/\eta)$ terms.

In practice, we may not know whether SH or SHB is more appropriate for our problem setting, since the budget threshold depends on unknown problem parameters, $\Delta,\sigma$. While it is reasonable to assume $\sigma$ is known\footnote{While this assumption is not needed for SH/SHB, we assume $\sigma^2$ is known here to isolate the key difficulty of the adaptive problem.} (e.g. by restricting rewards to be bounded, see \citet{bubeck_BAI}), $\Delta$ is still unknown.
To deal with this, we propose a regime adaptive method, \textbf{Adaptive Sequential Halving (SHA)}, which performs near-optimally regardless of the setting we are in. The main idea is to use some small portion of the budget to identify 
whether SH or SHB is more appropriate to play for the remaining budget.
In particular, we first sample the actions $0,1$ a total of $L<T$ times and use these samples to estimate the change in mean $\hat{\Delta}$. 
We use $\hat \Delta$ to estimate a budget threshold $\tau$. 
Then if the budget is smaller than $\tau$, we use SH for the remainder of the budget and if T is larger than the threshold $\tau$ we use SHB. 
SHA takes as input a parameter $\gamma>0$ and uses the estimated threshold
\begin{equation} \label{adaptive_threshold_T_definition_eqn}
    \tau = \gamma \frac{\sigma^2}{\hat{\Delta}^2} \log\left(\frac{1}{2\eta}\right).
\end{equation}
 SHA is written explicitly in 
 Appendix \ref{appendix: algo def}.

To run SHA we have to choose appropriate hyperparamaters $\gamma,L$. While we could pick $\gamma$ to match $T_1$ (from Theorems \ref{corr_large_budget_lb}, \ref{lower_bound_exp_form}), it turns out there are better choices both in theory and practice. 
In Theorem \ref{adaptive upper bound} we show that there exists a
universal
choice for $\gamma,L$ such that the SHA algorithm is near optimal for both small and large budgets simultaneously. 
Choices for $\gamma,L$ which perform well in practice are seen in Section \ref{section_experiments}.

\begin{theorem}\label{adaptive upper bound}
    Let $L=BT$ for some $B \in (2/T,1-2/T)$,
    $\gamma \in \left((\sqrt{104/B}+\sqrt{1.59})^2, 1800/(1-B)\right)$, 
    and define $\ell_\eta=\log _2 \left(1/2\eta\right)$.
    In an environment $v \in V(\Delta,\sigma)$, using SHA and with universal constants $c_1,c_2,c_3$
    $$
    \mathbb{P}(F_{T,v,\eta}) \leq
    \begin{cases}
        4\left\lceil \ell_\eta \right\rceil \exp\left(\frac{-c_1 (1-B) \Delta^2 T}{\sigma^2 \ell_\eta}\right), & T<T_1\\
        5\exp\left(-\frac{c_2 B(1-B) \Delta^2}{\sigma^2}T + c_3\ell_\eta\right), & T \geq T_1
    \end{cases}
    $$
\end{theorem}
The proof of Theorem \ref{adaptive upper bound} is given in Appendix \ref{Appendix_adaptive_algorithm}.
Theorem~\ref{adaptive upper bound} shows that SHA simultaneously matches the lower bounds in Theorems \ref{corr_large_budget_lb} and \ref{lower_bound_exp_form} for $T\geq T_1$ and $T<T_1$, up to constants and $\log\log (1/\eta)$ terms.

\begin{figure*}[t]
\centering
\begin{subfigure}{0.33\linewidth}
\centering
    \includegraphics[width=\linewidth]{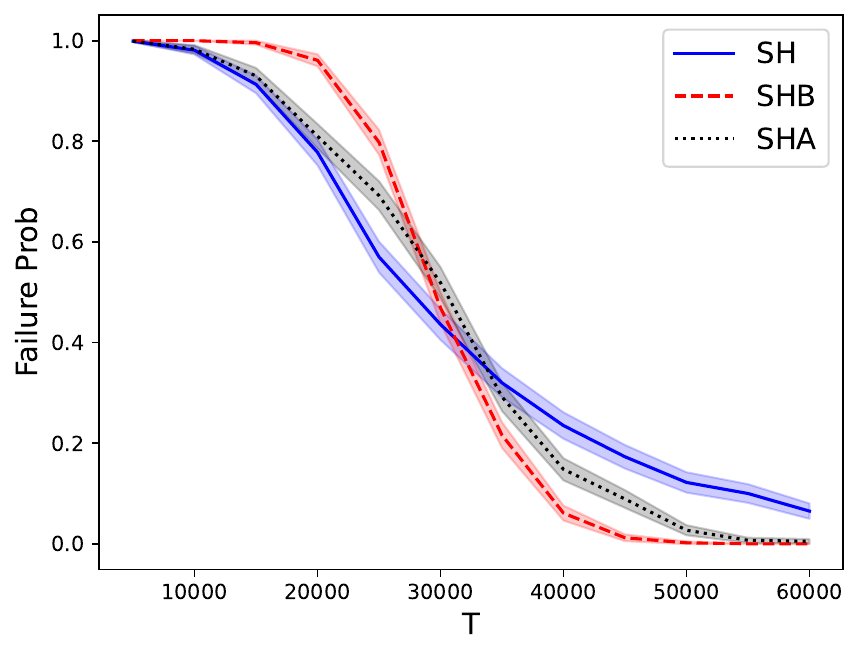}
        \caption{$\sigma=8, \eta=10^{-8}, x^*=0.7$}
    \label{fig:back_vs_noback1}
\end{subfigure}%
\begin{subfigure}{0.33\linewidth}
\centering
    \includegraphics[width=\linewidth]{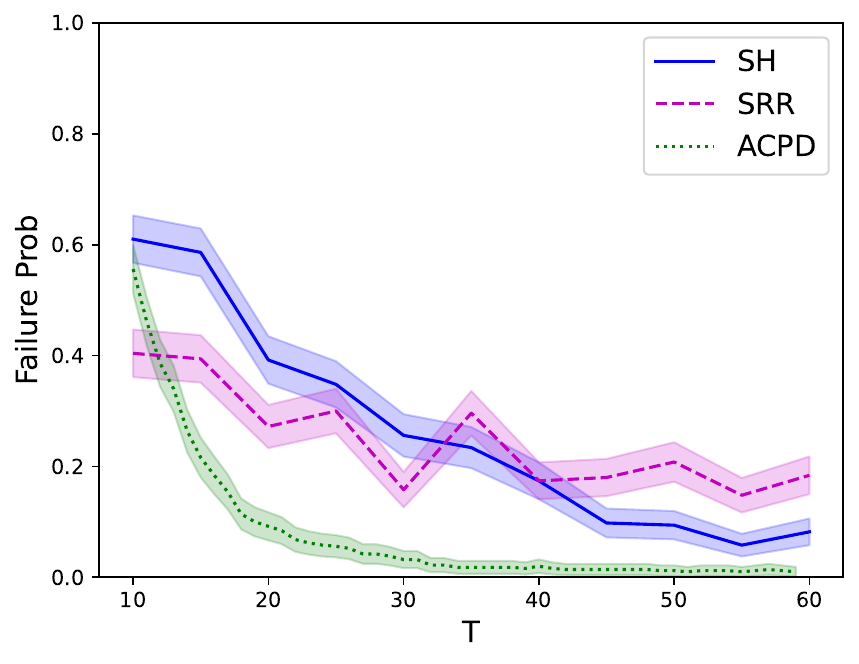}
        \caption{$\sigma=1,\eta=0.1,x^*=0.7$}
    \label{fig:all_algos1}
\end{subfigure}
\begin{subfigure}{0.33\linewidth}
\centering
    \includegraphics[width=\linewidth]{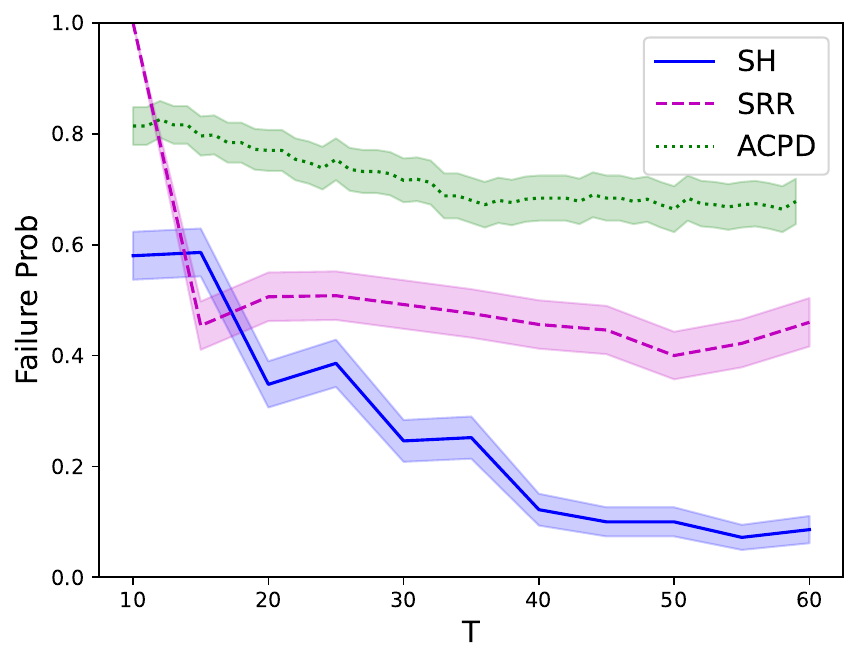}
        \caption{$\sigma=1, \eta=0.1,x^*=0.01$}
    \label{fig:all_algos2}
\end{subfigure}
\caption{Proportion of final estimates more than $\eta$ away from $x^*$  against the inputted budget, $T$, with Gaussian rewards, $\Delta=2$ and 90\% CIs. (a) compares the SH, SHB, and SHA by running each algorithm 1000 times with different budgets. (b,c) both compare SH, SRR, and ACPD. We run SH and SRR 500 times each at different budgets, while the anytime ACPD algorithm is run a total of 500 times for $T=60$ and we plot the evolution of ACPD's failure probability.}
\label{fig:3main_paper_figures}
\vskip -0.2in
\end{figure*}

\section{Experiments}\label{section_experiments} 
\paragraph{Comparing SH, SHB, and SHA} In Figure \ref{fig:back_vs_noback1} we demonstrate that SH or SHB can perform better than the other depending on the problem setting. 
We consider a synthetic environment with $\Delta=2,\sigma=8, x^*=0.7, \eta=10^{-8}$. Figure \ref{fig:back_vs_noback1} shows the error as a function of inputted budget. We see that SH has a smaller failure probability for smaller budgets. However, as the budget gets larger, SHB performs better and reaches near-zero probability faster. This supports our theoretical observations in Section \ref{section_regime_identification}. Similar results hold for other problem instances (see Appendix \ref{appendix:more sims}). We compare SHA to SH/SHB in the same environment. We chose the hyperparameters $L = T/20$ and $\gamma =120$ for SHA as these worked well across a variety of environments (see Appendix \ref{appendix:more sims}). We see from Figure \ref{fig:back_vs_noback1} that SHA can outperform SHB for small budgets and outperform SH for large budgets. Therefore, 
we observe that SHA performs well for all the budgets simultaneously.

{\bf Comparison with existing work} \quad We compare our proposed algorithms with existing methods. In Figures \ref{fig:all_algos1} and \ref{fig:all_algos2} we consider synthetic environments with $\Delta=2,\sigma=1, \eta=0.1$ only varying the location of the change point from $x^*=0.7$ to $x^*=0.01$, respectively.
We compare our SH algorithm to SRR \citep{active_Hall2003} and ACPD \citep{ACPD}. Due to the significant computational expense of repeated GP regression in the ACPD algorithm, we focus on smaller budget problems (where simulations are tractable) and consequently compare these algorithms to our small budget algorithm, SH. For ACPD, we use parameters suggested in \citet{ACPD} along with knowledge that rewards are Gaussian. We initialise ACPD with 10 random actions, and use GP-UCB \citep{GPUCB_paper} with Matérn kernel of smoothness $5/2$ for the Bayesian Optimisation. 
For the SRR algorithm
we use the proposed parameters in Section 4 of \citet{active_Hall2003}.
In particular we choose the number of actions in each stage $m=\log^{2+\beta} (T)$ and the CI radius parameter $\lambda=\log^{1+\alpha} (T)$, with $\alpha=1,\beta=0.1$. We show that tuning these parameters for particular environments does not lead to significant performance boosts 
in Appendix \ref{appendix:more sims}. There we also show that the method in \citet{Active_Lan_2009} (which is similar to SRR) also has very similar empirical performance to SRR. %

We first compare the performances of these algorithms when the change point is $x^*=0.7$ in Figure \ref{fig:all_algos1}. Here we see that the failure probability of ACPD decreases fastest while  SRR and SH are similar to each other, decreasing at a slower rate towards zero. However, when the change point is closer to the boundary ($x^*=0.01$) in Figure \ref{fig:all_algos2} we see that the performance of SRR and ACPD becomes significantly worse, whereas the performance of SH is relatively unchanged. This demonstrates that existing methods' performance is dependent on the change point being near the centre of the action space whereas our proposed policies perform well regardless of the location of $x^*$. Our methods are also simple, computationally inexpensive, and accompanied with tight non-asymptotic optimality guarantees. 

\section{Discussion}\label{section: conclusion}
In this paper, we studied the piecewise constant bandit problem and provided the first non-asymptotic problem dependent theoretical analysis of the problem. 
We developed two algorithms, SH and SHB, which achieve nearly minimax optimal error probabilities under different conditions on the budget. We then combined these two algorithms into a regime adaptive method SHA which is near optimal in both regimes simultaneously.
We complemented our theoretical results with simulations 
and provided a comparison to existing methods.

A natural extension of the piecewise constant bandit problem would be to allow for multiple abrupt changes in the reward function, or to allow the action space to be multi-dimensional. Similarly, it would be interesting to extend our methods to the piecewise smooth setting as in \citet{active_Hall2003}.
We expect significant innovation to be required to extend our non-asymptotic results to these settings, although we hope our methods will provide a useful starting point.   
The findings in this paper already represent a significant advancement in understanding the complexity of 
choosing samples to learn the location of a
change point in unknown, noisy environments. 

\bibliography{bibliography.bib}

\begin{thebibliography}{33}
\providecommand{\natexlab}[1]{#1}
\providecommand{\url}[1]{\texttt{#1}}
\expandafter\ifx\csname urlstyle\endcsname\relax
  \providecommand{\doi}[1]{doi: #1}\else
  \providecommand{\doi}{doi: \begingroup \urlstyle{rm}\Url}\fi

\bibitem[Abbasi-Yadkori et~al.(2011)Abbasi-Yadkori, P\'{a}l, and Szepesv\'{a}ri]{linear_bandits_yadkori}
Y.~Abbasi-Yadkori, D.~P\'{a}l, and C.~Szepesv\'{a}ri.
\newblock Improved algorithms for linear stochastic bandits.
\newblock In \emph{Advances in Neural Information Processing Systems}, 2011.

\bibitem[Agarwal et~al.(2011)Agarwal, Foster, Hsu, Kakade, and Rakhlin]{convex_bandits}
A.~Agarwal, D.~P. Foster, D.~J. Hsu, S.~M. Kakade, and A.~Rakhlin.
\newblock Stochastic convex optimization with bandit feedback.
\newblock In \emph{Advances in Neural Information Processing Systems}, 2011.

\bibitem[Aminikhanghahi and Cook(2017)]{CP_survey}
S.~Aminikhanghahi and D.~Cook.
\newblock A survey of methods for time series change point detection.
\newblock \emph{Knowledge and Information Systems}, 2017.

\bibitem[Audibert et~al.(2010)Audibert, Bubeck, and Munos]{bubeck_BAI}
J.-Y. Audibert, S.~Bubeck, and R.~Munos.
\newblock Best arm identification in multi-armed bandits.
\newblock In \emph{COLT 2010 - The 23rd Conference on Learning Theory}, 2010.

\bibitem[Brown et~al.(2001)Brown, Cai, and DasGupta]{simple_gaussian_CI}
L.~D. Brown, T.~T. Cai, and A.~DasGupta.
\newblock {Interval Estimation for a Binomial Proportion}.
\newblock \emph{Statistical Science}, 2001.

\bibitem[Bubeck et~al.(2011)Bubeck, Munos, Stoltz, and Szepesvári]{treeAlgo}
S.~Bubeck, R.~Munos, G.~Stoltz, and C.~Szepesvári.
\newblock X-armed bandits.
\newblock \emph{Journal of Machine Learning Research}, 2011.

\bibitem[Carpentier and Locatelli(2016)]{Carpentier2016TightLowerBounds}
A.~Carpentier and A.~Locatelli.
\newblock Tight (lower) bounds for the fixed budget best arm identification bandit problem.
\newblock In \emph{Annual Conference Computational Learning Theory}, 2016.

\bibitem[Castro and Nowak(2008)]{Acitve_Castro2008}
R.~M. Castro and R.~D. Nowak.
\newblock Minimax bounds for active learning.
\newblock \emph{IEEE Transactions on Information Theory}, 2008.

\bibitem[Castro et~al.(2005)Castro, Willett, and Nowak]{Active_Castro2005}
R.~M. Castro, R.~M. Willett, and R.~D. Nowak.
\newblock Faster rates in regression via active learning.
\newblock In \emph{Advances in Neural Information Processing Systems}, 2005.

\bibitem[Chen and Gupta(2012)]{ChenGuptaCP_Book}
J.~Chen and A.~K. Gupta.
\newblock \emph{{Parametric Statistical Change Point Analysis: With Applications to Genetics, Medicine, and Finance; 2nd ed.}}
\newblock Springer, 2012.

\bibitem[Cheshire et~al.(2021)Cheshire, Menard, and Carpentier]{problem_dependent_threshold}
J.~Cheshire, P.~Menard, and A.~Carpentier.
\newblock Problem dependent view on structured thresholding bandit problems.
\newblock In \emph{Proceedings of the 38th International Conference on Machine Learning}, 2021.

\bibitem[Dasgupta(2005)]{discontinuity_paper_active_learning}
S.~Dasgupta.
\newblock Coarse sample complexity bounds for active learning.
\newblock \emph{Advances in Neural Information Processing Systems}, 2005.

\bibitem[Garivier and Moulines(2011)]{Discounted_UCB_non_stationary}
A.~Garivier and E.~Moulines.
\newblock On upper-confidence bound policies for switching bandit problems.
\newblock In \emph{Algorithmic Learning Theory}, 2011.

\bibitem[Garivier et~al.(2016)Garivier, Ménard, and Stoltz]{Garivier_chain_rule_trick}
A.~Garivier, P.~Ménard, and G.~Stoltz.
\newblock Explore first, exploit next: The true shape of regret in bandit problems.
\newblock \emph{Mathematics of Operations Research}, 2016.

\bibitem[Gramacy and Lee(2008)]{AdaptiveDesignSupercomputer}
R.~Gramacy and H.~Lee.
\newblock Adaptive design and analysis of supercomputer experiments.
\newblock \emph{Technometrics}, 2008.

\bibitem[Hall and Molchanov(2003)]{active_Hall2003}
P.~Hall and I.~Molchanov.
\newblock {Sequential methods for design-adaptive estimation of discontinuities in regression curves and surfaces}.
\newblock \emph{The Annals of Statistics}, 2003.

\bibitem[Hayashi et~al.(2019)Hayashi, Kawahara, and Kashima]{ACPD}
S.~Hayashi, Y.~Kawahara, and H.~Kashima.
\newblock Active change-point detection.
\newblock In \emph{Proceedings of The Eleventh Asian Conference on Machine Learning}, 2019.

\bibitem[Karnin et~al.(2013)Karnin, Koren, and Somekh]{sequential_halving_original}
Z.~Karnin, T.~Koren, and O.~Somekh.
\newblock Almost optimal exploration in multi-armed bandits.
\newblock In \emph{Proceedings of the 30th International Conference on Machine Learning}, 2013.

\bibitem[Karp and Kleinberg(2007)]{karp_kleinberg}
R.~M. Karp and R.~D. Kleinberg.
\newblock Noisy binary search and its applications.
\newblock In \emph{ACM-SIAM Symposium on Discrete Algorithms}, 2007.

\bibitem[Kleinberg et~al.(2008)Kleinberg, Slivkins, and Upfal]{ZoomingAlgo}
R.~Kleinberg, A.~Slivkins, and E.~Upfal.
\newblock Multi-armed bandits in metric spaces.
\newblock In \emph{Proceedings of the Fortieth Annual ACM Symposium on Theory of Computing}, 2008.

\bibitem[Lan et~al.(2009)Lan, Banerjee, and Michailidis]{Active_Lan_2009}
Y.~Lan, M.~Banerjee, and G.~Michailidis.
\newblock Change-point estimation under adaptive sampling.
\newblock \emph{The Annals of Statistics}, 2009.

\bibitem[Lattimore and Szepesvári(2020)]{BanditAlgosBook}
T.~Lattimore and C.~Szepesvári.
\newblock \emph{Bandit Algorithms}.
\newblock Cambridge University Press, 2020.

\bibitem[Locatelli et~al.(2016)Locatelli, Gutzeit, and Carpentier]{locatelli16_thresholding}
A.~Locatelli, M.~Gutzeit, and A.~Carpentier.
\newblock An optimal algorithm for the thresholding bandit problem.
\newblock In \emph{Proceedings of The 33rd International Conference on Machine Learning}, 2016.

\bibitem[Lu et~al.(2020)Lu, Banerjee, and Michailidis]{fast_offline_cp}
Z.~Lu, M.~Banerjee, and G.~Michailidis.
\newblock Intelligent sampling for multiple change-points in exceedingly long time series with rate guarantees.
\newblock In \emph{ArXiv Preprint ArXiv:1710.07420}, 2020.

\bibitem[Mannor et~al.(2004)Mannor, Tsitsiklis, Bennett, and Cesa-bianchi]{MannorLowerBoundPaper}
S.~Mannor, J.~Tsitsiklis, K.~Bennett, and N.~Cesa-bianchi.
\newblock The sample complexity of exploration in the multi-armed bandit problem.
\newblock 2004.

\bibitem[Park et~al.(2021)Park, Qiu, Carpena-Núñez, Rao, Susner, and Maruyama]{park2021sequentialAdaptiveDesignForJumpRegression}
C.~Park, P.~Qiu, J.~Carpena-Núñez, R.~Rao, M.~Susner, and B.~Maruyama.
\newblock Sequential adaptive design for jump regression estimation.
\newblock In \emph{ArXiv Preprint ArXiv:1904.01648}, 2021.

\bibitem[Park et~al.(2023)Park, Waelder, Kang, Maruyama, Hong, and Gramacy]{ActiveLearningPiecewiseGP}
C.~Park, R.~Waelder, B.~Kang, B.~Maruyama, S.~Hong, and R.~Gramacy.
\newblock Active learning of piecewise gaussian process surrogates.
\newblock In \emph{ArXiv Preprint ArXiv:2301.08789}, 2023.

\bibitem[Sikorski(1982)]{bisection_optimal}
K.~Sikorski.
\newblock Bisection is optimal.
\newblock \emph{Numerische Mathematik}, 40:\penalty0 111--118, 1982.

\bibitem[Srinivas et~al.(2010)Srinivas, Krause, Kakade, and Seeger]{GPUCB_paper}
N.~Srinivas, A.~Krause, S.~Kakade, and M.~Seeger.
\newblock Gaussian process optimization in the bandit setting: No regret and experimental design.
\newblock In \emph{ICML 2010 - Proceedings, 27th International Conference on Machine Learning}, 2010.

\bibitem[Thuot et~al.(2024)Thuot, Carpentier, Giraud, and Verzelen]{thuot2024activeclusteringbanditfeedback}
V.~Thuot, A.~Carpentier, C.~Giraud, and N.~Verzelen.
\newblock Active clustering with bandit feedback.
\newblock In \emph{ArXiv Preprint ArXiv:2406.11485}, 2024.

\bibitem[Wainwright(2019)]{wainwright_2019}
M.~J. Wainwright.
\newblock \emph{High-Dimensional Statistics: A Non-Asymptotic Viewpoint}.
\newblock Cambridge Series in Statistical and Probabilistic Mathematics. Cambridge University Press, 2019.

\bibitem[Yang et~al.(2022)Yang, Zhong, and Tan]{Yang2022OptimalClustering}
J.~Yang, Z.~Zhong, and V.~Y.~F. Tan.
\newblock Optimal clustering with bandit feedback.
\newblock \emph{Journal of Machine Learning Research}, 2022.

\bibitem[Yavas et~al.(2024)Yavas, Huang, Tan, and Scarlett]{yavas2024generalframeworkclusteringdistribution}
R.~C. Yavas, Y.~Huang, V.~Y.~F. Tan, and J.~Scarlett.
\newblock A general framework for clustering and distribution matching with bandit feedback.
\newblock In \emph{ArXiv Preprint ArXiv:2409.05072}, 2024.

\end{thebibliography}

 \newpage
 \onecolumn

\appendix
\section*{APPENDIX}
\tableofcontents

\newpage

\section{Explicit Algorithms} \label{appendix: algo def}
We explicitly write the SH, SHB and SHA algorithms below.

\begin{algorithm}[H]
\caption{Sequential halving with backtracking (SHB)}\label{alg:backtracking}
\begin{algorithmic}[1]
\State {\bfseries Input:} $\eta \in (0,1/2),$ and budget $T$
\State $\mathcal{A}^1 = \{0,a_1^1,a_2^1,a_3^1,1\} \gets \{0,0,1/2,1,1\}$
\State $J \gets \lceil 6\log  (1/2\eta) \rceil$
\State $t_j \gets \lfloor \frac{T}{5J}\rfloor$
\For{phase $j$ in $1, ... , J$}
\State Play each action in $\mathcal{A}^j = \{0,a_1^j,a_2^j,a_3^j,1$\}, \, $t_j$ times
\If{$E_{P,j}$ from \eqref{event_E_P} holds}
\State $\mathcal{A}^{j+1} \leftarrow P(\mathcal{A}^{j})$ \Comment{Backtrack}
\ElsIf{$E_{R,j}$ from \eqref{event_E_R} holds}
    \State $\mathcal{A}^{j+1} \leftarrow R(\mathcal{A}^{j})$ \Comment{Zoom in to the right}
\ElsIf{$E_{L,j}$ from \eqref{event_E_L} holds}
    \State $\mathcal{A}^{j+1} \leftarrow L(\mathcal{A}^{j})$ \Comment{Zoom in to the left}
\EndIf
\EndFor
\State \textbf{Return:} $\hat{x}_T = a_2^{J+1}$
\end{algorithmic}
\end{algorithm}

\begin{algorithm}[H]
\caption{Sequential halving without backtracking (SH)}\label{alg:fixedbudget}
\begin{algorithmic}
\State \textbf{Input:} $\eta \in (0,1/2),$ and budget $T$
\State $\{a_1^1,a_2^1,a_3^1\} \gets \{0,1/2,1\}$
\State $J \gets \lceil \log _2 (1/2\eta) \rceil$
\State $t_j \gets \lfloor \frac{T}{3J}\rfloor$
\For{phase $j$ in $1, ... , J$}
\State Play each action $a_1^j,a_2^j,a_3^j$, \, $t_j$ times 
\If{$E_{R,j}$ from \eqref{event_E_R} holds}
    \State $\{a_1^{j+1},a_2^{j+1},a_3^{j+1}\} \leftarrow  \{a_2^j,(a_2^j+a_3^j)/2,a_3^j\}$ \Comment{Zoom in to the right}
\ElsIf{$E_{L,j}$ from \eqref{event_E_L} holds}
    \State $\{a_1^{j+1},a_2^{j+1},a_3^{j+1}\} \leftarrow \{a_1^j,(a_1^j+a_2^j)/2,a_2^j\}$ \Comment{Zoom in to the left}
\EndIf
\EndFor
\State \textbf{Return} $\hat{x}_T = a_2^J$
\end{algorithmic}
\end{algorithm}

\begin{algorithm}[H] 
\caption{Adaptive Sequential Halving (SHA)}\label{alg:adaptive}
\begin{algorithmic}[1]
\State {{\bfseries Input:}} $\gamma>0,L \geq 2$ \Comment{Hyperparameters}
\State {\bfseries Input:} $\eta \in (0,1/2),\sigma>0$ and budget $T$
\State Play actions 0,1 each $L/2$ times
\Comment{Initial phase to estimate threshold}
\State Calculate $\hat{\Delta}_L = |\hat{\mu}_{1,L} - \hat{\mu}_{0,L}|$
\State Calculate $\tau = \gamma \frac{\sigma^2}{\hat{\Delta}^2_L} \log\left(\frac{1}{2\eta}\right)$
\Comment{Threshold $\tau$ from \eqref{adaptive_threshold_T_definition_eqn}}
\If{$T \geq \tau$} 
\State $\hat{x}_{T-L}^{SHB} \gets SHB(T-L, \eta)$
\Comment{Play SHB for remainder of budget}
\State \textbf{Return:} $\hat{x}_T^{SHA} = \hat{x}_{T-L}^{SHB}$
\EndIf
\If{$T < \tau$}
\State $\hat{x}_{T-L}^{SH} \gets SH(T-L, \eta)$
\Comment{Play SH for remainder of budget}
\State \textbf{Return:} $\hat{x}_T^{SHA} = \hat{x}_{T-L}^{SH}$
\EndIf
\end{algorithmic}
\end{algorithm}

\newpage
\section{Additional Experiments} \label{appendix:more sims}

\subsection{Further Empirical Comparisons of SH, SHB and SHA with Other $\eta$ Values} \label{appendix_experiments_SHA}
In Figure \ref{fig:back_vs_noback1} from Section \ref{section_experiments}, we demonstrate that when $\eta$ is very small ($\eta=10^{-8}$ in that case) the SH algorithm will outperform SHB for smaller budgets, whereas the SHB algorithm will reach near zero failure probabilities faster than SH. This matches our discussion of our theoretical results in Section \ref{section_regime_identification}. 
This also matches our intuition.
Namely, in order for SH to return a good estimate, it must eliminate the correct half of the action space in \emph{all} of the $\log_2 (1/2\eta)$ phases. This occurs with very small probability when $\eta$ is small (i.e. the number of phases is large) and therefore we can benefit from using SHB which explores more and only requires correct decisions in some portion of the phases for a good estimate.
In Figure \ref{fig:back_vs_noback1} we also illustrate the desired effect of SHA, which performs well for both large and small budgets simultaneously.\\

\begin{figure}[H]
\centering
\begin{subfigure}{0.49\linewidth}
\centering
    \includegraphics[width=\linewidth]{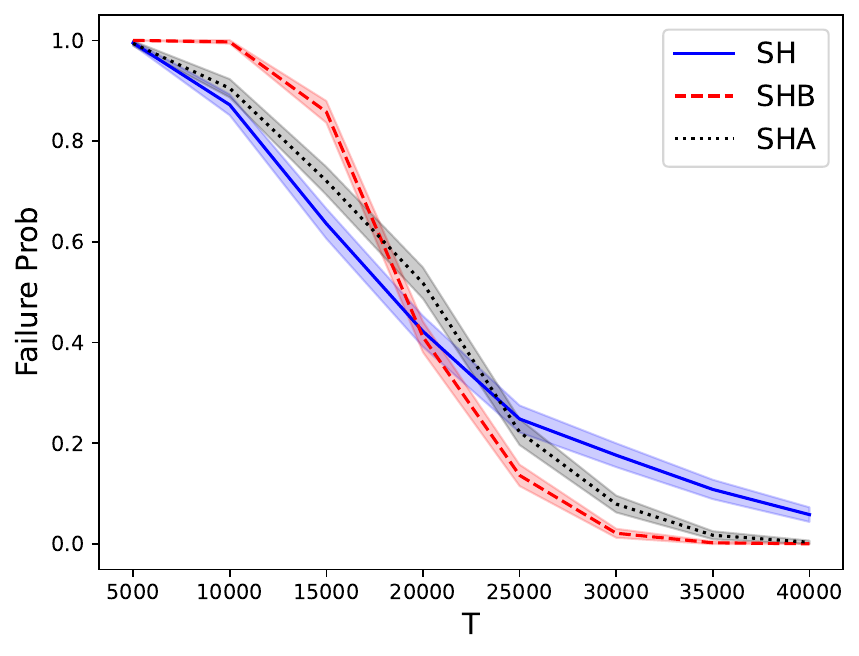}
        \caption{$\sigma=7, \eta=10^{-7}, x^*=0.7$}
    \label{fig:adaptive_eta_7}
\end{subfigure}%
\begin{subfigure}{0.49\linewidth}
\centering
    \includegraphics[width=\linewidth]{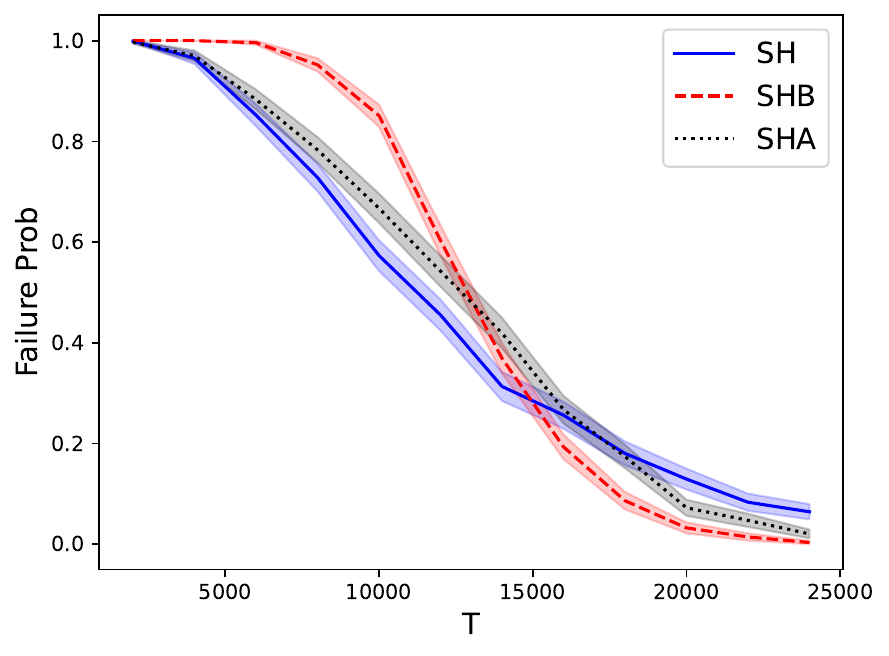}
        \caption{$\sigma=2.5^2, \eta=10^{-6}, x^*=0.7$}
    \label{fig:adaptive_eta_6}
\end{subfigure}
\caption{Proportion of final estimates more than $\eta$ away from $x^*$  against the inputted budget, $T$, with Gaussian rewards, $\Delta=2$ and 90\% CIs. SH, SHB, and SHA were each run 1000 times with different budgets.}
\label{fig:adaptive_small_eta}
\end{figure}

\begin{figure}[H]
\centering
\begin{subfigure}{0.49\linewidth}
\centering
    \includegraphics[width=\linewidth]{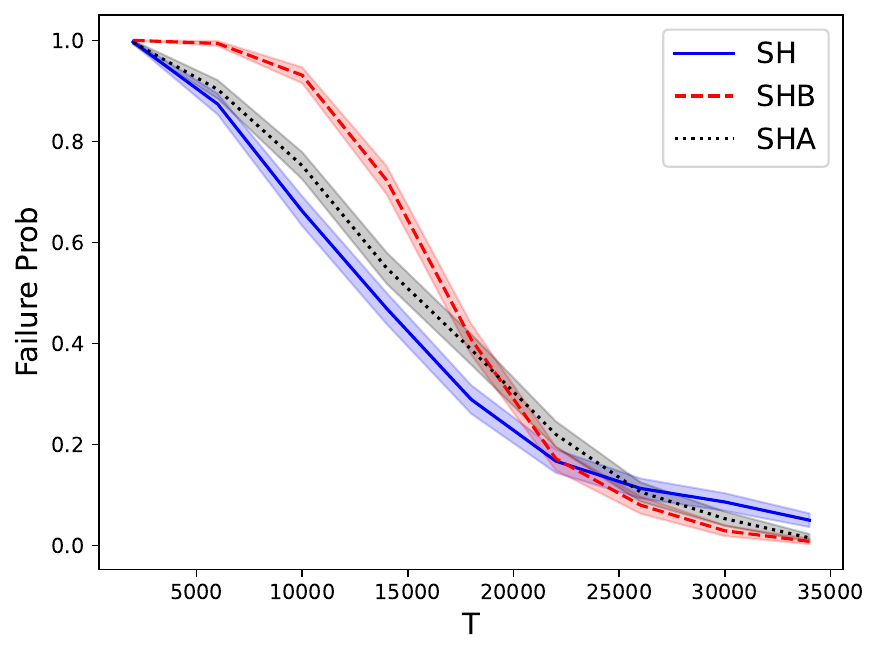}
        \caption{$\sigma=9, \eta=10^{-4}, x^*=0.7$}
    \label{fig:adaptive_eta_4}
\end{subfigure}%
\begin{subfigure}{0.49\linewidth}
\centering
    \includegraphics[width=\linewidth]{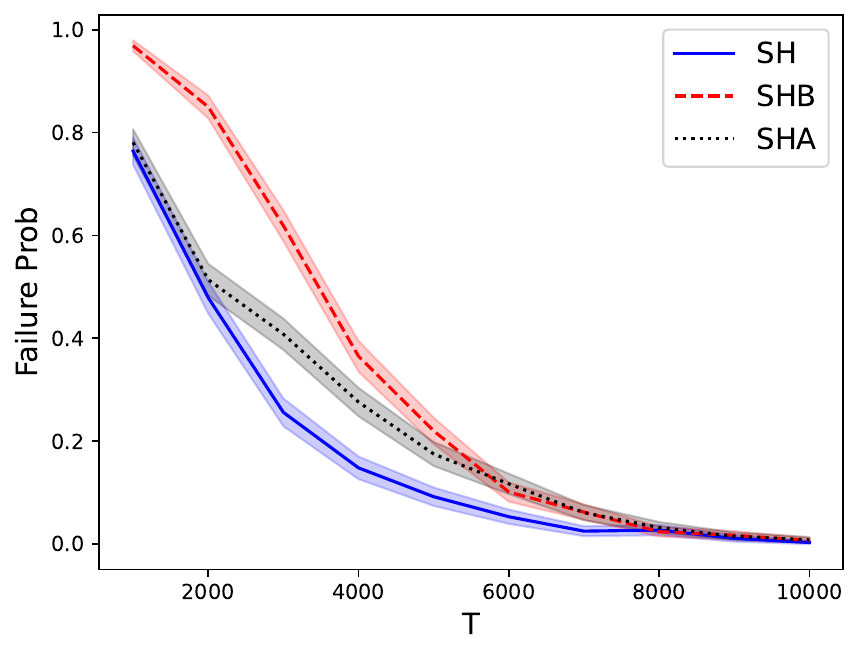}
        \caption{$\sigma=2.5^2, \eta=10^{-2}, x^*=0.7$}
    \label{fig:adaptive_eta_2}
\end{subfigure}
\caption{Proportion of final estimates more than $\eta$ away from $x^*$  against the inputted budget, $T$, with Gaussian rewards, $\Delta=2$ and 90\% CIs. SH, SHB, and SHA were each run 1000 times with different budgets.}
\label{fig:adaptive_big_eta}
\end{figure}

In Figure \ref{fig:adaptive_small_eta} we now consider two other settings to compare SH, SHB, and SHA where our simulations provide similar conclusions. In particular, in Figures \ref{fig:adaptive_eta_7} and \ref{fig:adaptive_eta_6} we consider settings with $\eta=10^{-7}$ and $10^{-6}$ respectively. In both cases, we once again see that for smaller budgets SH outperforms SHB, while SHB is able to reach near-zero failure probabilities faster than SH in larger budgets. Hence, SH is still more appropriate to use for settings with smaller budgets whereas SHB is more appropriate to use for settings with larger budgets. Additionally we show that SHA, with the hyperparameters $L = T/20$ and $\gamma =120$ performs well in both of these additional settings in Figures \ref{fig:adaptive_eta_7} and \ref{fig:adaptive_eta_6}. In particular, for both of these new settings, SHA outperforms SHB for smaller budgets and outperforms SH for larger budgets. We conclude SHA performs well for all budgets simultaneously. We note that the hyperparameters $L = T/20$ and $\gamma =120$ are the same as those used for Figure \ref{fig:back_vs_noback1}, suggesting that this is a good choice of hyperparameters for SHA that work well in a variey of settings. (This will be further supported by Figure \ref{fig:adaptive_big_eta}.)\\

In Figure \ref{fig:adaptive_big_eta}, we also compare SH, SHB, and SHA in settings with larger choices of $\eta$. In particular we consider $\eta=10^{-4}$ and $\eta=10^{-2}$ in Figures \ref{fig:adaptive_eta_4} and \ref{fig:adaptive_eta_2}, respectively. From these figures, we firstly note that our adaptive algorithm SHA, with the hyperparameters $L = T/20$ and $\gamma =120$, still performs well regardless of the budget. Secondly, from Figures \ref{fig:adaptive_eta_4} and \ref{fig:adaptive_eta_2} (as well as \ref{fig:adaptive_eta_7} and \ref{fig:adaptive_eta_6}), we observe the interesting phenomenon that the advantages of using SHB over SH become less significant as $\eta$ becomes larger. This matches our intuition. In particular, for SH to return a good estimate, it must make the correct decision in all $\log_2(1/2\eta)$ phases. This can occur with high probability when $\eta$ is large (i.e. the number of phases is small). Therefore the exploitative nature of SH in settings with larger $\eta$ can outweigh the benefits from the additional exploration in SHB (which can allow mistakes in some phases).
These observations also match our theoretical results. In particular for larger values of $\eta$, the the $\frac{T\Delta^2}{36\sigma^2 \log_2(1/2\eta)}$ term from our guarantee for SH (Theorem \ref{fixed budget upper bound}) can be similar to or larger than the $\frac{\Delta^2}{600\sigma^2}T$ term from our guarantee for SHB (Theorem \ref{fixed budget upper bound backtracking}). In which case our guarantee for SH would be similar to or better than SHB.

\subsection{Comparison of All Algorithms}

\begin{figure}[H]
\centering
\begin{subfigure}{0.49\linewidth}
\centering
    \includegraphics[width=\linewidth]{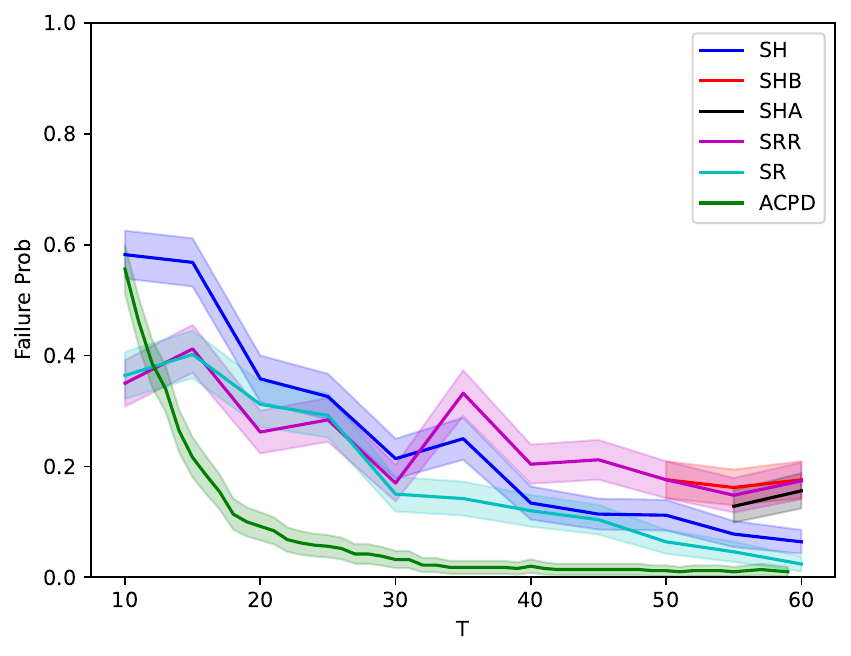}
        \caption{$\sigma=1,\eta=0.1,x^*=0.7$}
    \label{fig:app_all_algos_1}
\end{subfigure}%
\begin{subfigure}{0.49\linewidth}
\centering
    \includegraphics[width=\linewidth]{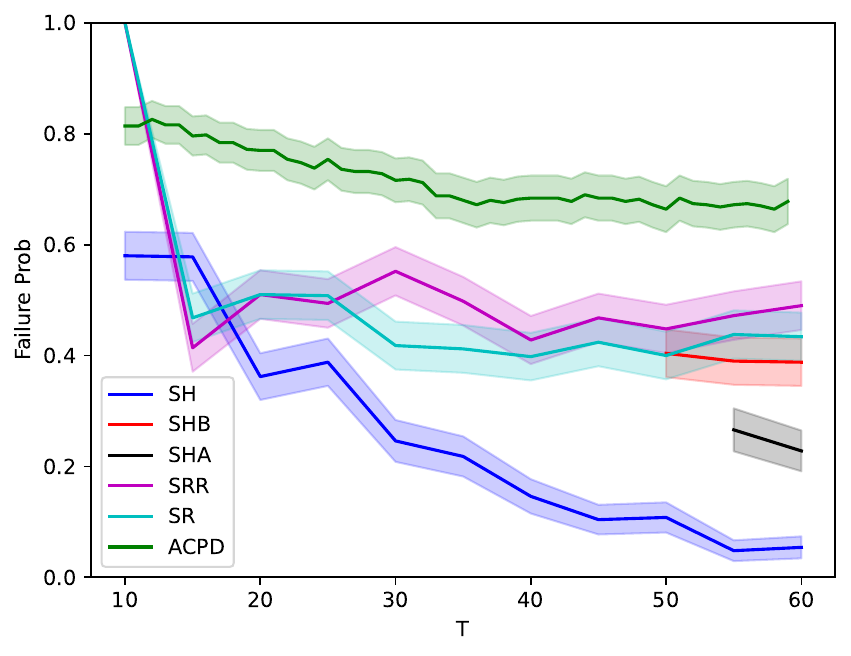}
        \caption{$\sigma=1,\eta=0.1,x^*=0.01$}
    \label{fig:app_all_algos_2}
\end{subfigure}
\caption{Proportion of final estimates more than $\eta$ away from $x^*$  against the inputted budget, $T$, with Gaussian rewards, $\Delta=2$ and 90\% CIs. SH, SHB, SHA, SRR, and SR were each run 500 times with different budgets, while the anytime ACPD algorithm is run a total of 500 times for $T=60$ and we plot the evolution of ACPD's failure probability.}
\label{fig:all_algos_appendix}
\end{figure}

We include the additional algorithms of SHB, SHA and SR \citep{Active_Lan_2009} to to the setting of Figures \ref{fig:all_algos1} and \ref{fig:all_algos2} in Figures \ref{fig:app_all_algos_1} and \ref{fig:app_all_algos_2} here. Recall from Section \ref{section_experiments} that, due to the very computationally expensive existing methods (e.g. ACPD), we consider settings with very limited budgets. Our small budget algorithm SH is much better suited to these very limited budgets and hence this was the main algorithm we compared with existing work. Furthermore, in this setting SHB (and similarly SHA) requires a budget $T\geq 50$ in order for there to be at least one action for each sampling point in each phase. We plot the performance of SHB/SHA for such budgets in Figures \ref{fig:app_all_algos_1} and \ref{fig:app_all_algos_2}. 
More thorough simulations comparing SH, SHB, and SHA can be seen in Appendix \ref{appendix_experiments_SHA}.

\paragraph{SRR Algorithm and Tuning}
In the Sequential Refinement with Reassessment Algorithm (SRR), \citet{active_Hall2003} propose a multi-stage method where they first spend half of the budget uniformly exploring the space and they then spend every subsequent stage sampling $m$ actions evenly across a confidence interval for the change point constructed in the previous stage. The width of this confidence interval is influenced by parameter $\lambda$. In each stage they also have a reassessment criteria in which they test for the presence of a change point during the current stage (within the current CI for the change point) and if it is not significant with level $1-\epsilon$ confidence, they ``reassess" and return to the previous confidence interval constructed. In Section 4 of \citet{active_Hall2003}, the authors propose setting $m=\log^{2+\beta} (T)$ and $\lambda=\log^{1+\alpha} (T)$ with $0<\alpha<1+\beta$ with $\beta>0$. Since we would like for there to be at least two stages and to satisfy these conditions, for experiments in Figures \ref{fig:all_algos1} and \ref{fig:all_algos2} (similarly \ref{fig:app_all_algos_1}, \ref{fig:app_all_algos_2}) we set $\beta=0.1,\alpha=1$. 

\begin{figure}[H]
\centering
\begin{subfigure}{0.49\linewidth}
\centering
    \includegraphics[width=\linewidth]{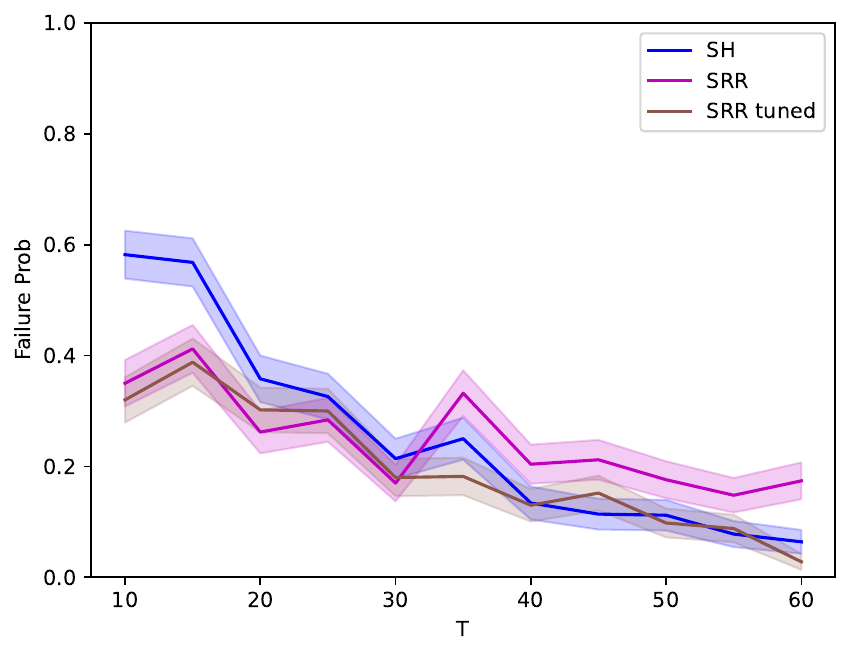}
        \caption{$\sigma=1,\eta=0.1,x^*=0.7$}
    \label{fig:app_tuned_plots_1}
\end{subfigure}%
\begin{subfigure}{0.49\linewidth}
\centering
    \includegraphics[width=\linewidth]{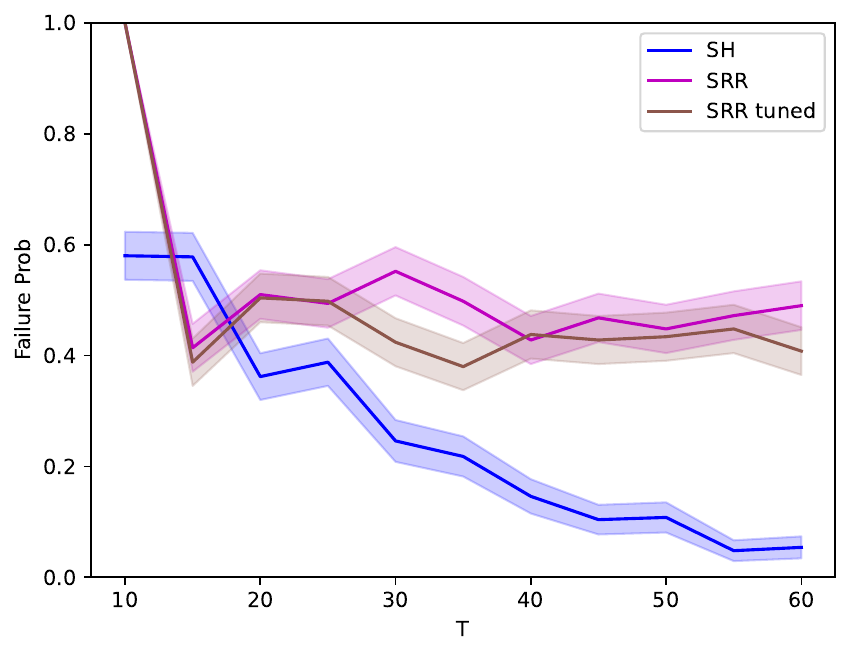}
        \caption{$\sigma=1,\eta=0.1,x^*=0.01$}
    \label{fig:app_tuned_plots_2}
\end{subfigure}
\caption{Proportion of final estimates more than $\eta$ away from $x^*$  against the inputted budget, $T$, with Gaussian rewards, $\Delta=2$ and 90\% CIs. SH, and tuned SRR were each run 500 times with different budgets. In (a) the tuned constants for SRR tuned were $\alpha \approx 0.28, \beta \approx 0.41$. In (b) the tuned constants were $\alpha \approx 1.29, \beta \approx 0.29$.}
\label{fig:appendix_SRR_tuning}
\vspace{5in}
\end{figure}

One might be concerned that the negative performance by SRR compared to SH when the change point is near the boundary in Figure \ref{fig:all_algos2}, might be due to poorly chosen parameters $\alpha,\beta$. In practice, we will not know which parameters perform best in which environment. However, we can show that tuning these parameters does not significantly improve the performance in these settings. To do so, we first note that $\beta$ should be set to be at most $0.4$ since otherwise this would mean for $T\leq 60$ that there is only one stage. With budget $T=60$ for 10 values of $\beta$ (between 0 and 0.4), and at each value for $\beta$ we run 5 values for $\alpha$ (between 0 and $1+\beta$) 500 times. Across these 50 different $\alpha,\beta$ pairs chosen, we select the pair with minimal failure probability. We did this tuning for both environments studied in Figures \ref{fig:all_algos1} and \ref{fig:all_algos2} individually. For each we then plot the performance of SRR with these tuned parameters against the SH algorithm in Figures \ref{fig:app_tuned_plots_1}, \ref{fig:app_tuned_plots_2}.

\paragraph{SR Algorithm}
We refer to the algorithm proposed by \citet{Active_Lan_2009} as Sequential Refinement (SR). This is because the algorithm itself is extremely similar to SRR, except the policy never reassess once it has ``zoomed in" to a particular region of the action space. (Furthermore, they do not spend half of the budget initially exploring the space.) In particular, the SR algorithm begins by splitting the budget into $L$ stages. Then, in each stage SR plays actions evenly across a confidence interval for the change point (constructed using samples in the previous stage). For their multi-stage method \citet{Active_Lan_2009} generally recommend splitting the budget evenly into $L$ stages such that in each phase SR plays around $30-50$ actions. Hence, for these very limited budget settings with $T\leq 60$ seen in Figures \ref{fig:all_algos1} and \ref{fig:all_algos2}, we choose the smallest number of stages $L=2$. For the construction of the confidence intervals we use the proposed form shown in equation (10) of \citet{Active_Lan_2009}, which requires knowledge of the signal to noise ratio $\Delta/\sigma$. Note that we would not have access to this in practice. We plot the performance of SR in settings described in Figures \ref{fig:app_all_algos_1} and \ref{fig:app_all_algos_2}. Here we see that the the failure probability when running SR is very similar to SRR regardless of the position of the change point ($x^*=0.7$ or $x^*=0.01$). Furthermore, SH still significantly outperforms SR when the change is near the boundary $x^*=0.01$.

\subsection{Confidence intervals}
Confidence intervals in all plots for the failure probabilities of different algorithms were calculated using a simple Gaussian approximation, as seen in equation (1) of \citep{simple_gaussian_CI}. This was done using the set of all results at each budget for each algorithm.

\newpage
\section*{Proofs For Large Budgets}

\section{Proofs For Large Budget Upper Bound} \label{app_largeT_upper_proofs}

\subsection{Proof for Theorem \ref{fixed budget upper bound backtracking}}
\begin{proof}
Note that, since Algorithm \ref{alg:backtracking} is an extension of \citet{problem_dependent_threshold}, proofs for some technical lemmas (e.g. Lemmas \ref{lemma_ub_num_errors}, \ref{lemma_prob_num_errors}) are similar to existing work, up to the inclusion of an infinite action space and different elimination/backtracking probabilities. We write the main steps here and prove lemmas below. %

Firstly, for all $j$ in $\{1,...,J\}$, let $G_j$ be the ``good event'' 
\begin{align}
    G_j = &\left\{E_{P,j} \cap \left\{x^* \in (0,1) \backslash [a_1^j,a_3^j)\right\}\right\}
    \cup \left\{  E_{P,j}^C \cap E_{R,j} \cap \left\{x^* \in [a_2^j,a_3^j) \right\}  \right\} \label{eq_Gj_def}\\  
    &\cup \left\{   E_{P,j}^C \cap E_{L,j} \cap\left\{x^* \in  [a_1^j,a_2^j)\right\} \right\} \nonumber  
\end{align}
in which we make the correct decision in eliminating or backtracking in phase $j$. Namely, when the change point is not within the remaining action space in phase $j$,  $x^* \in (0,1) \backslash [a_1^j,a_3^j)$, we correctly backtrack. When $x^* \in [a_2^j,a_3^j)$ we correctly zoom into the right and eliminate the left half of region. And when $x^* \in  [a_1^j,a_2^j)$ we correctly zoom into the left and eliminate the right half of the remaining action space, see Figure \ref{fig_example_phases} again for illustration of the regions.

The first thing to note is that we need not make the correct decision in every phase. In particular, we can incorrectly eliminate a region when it contains the change point or incorrectly backtrack when the remaining action space contains the change point a limited number of times. As long as we fail in less than 1/4 of the phases, we will still achieve our objective $|\hat{x}_T-x^*|<\eta$. This is shown in Lemma~\ref{lemma_ub_num_errors}.

\begin{lemma} \label{lemma_ub_num_errors}
    Let $\eta < 1/4$. Under Algorithm \ref{alg:backtracking}, if we have $\sum_{j=1}^J \mathds{1}\{G_j^C\} < J/4$, then our final estimate will be within $\eta$ of the true change point, $|\hat{x}_T-x^*|<\eta$.
\end{lemma}

We then calculate a lower bound on the probability of making the correct decision in round $j$, given the actions and rewards from all previous phases $1,...,j-1$.

\begin{lemma} \label{lemma_ub_phase_j}
    Let $\mathcal{F}_{j-1}$ be the sigma algebra generated by all actions and rewards in the first $j-1$ phases.  Under Algorithm \ref{alg:backtracking} we have %
    $$\mathbb{P}(G_j|\mathcal{F}_{j-1}) \geq 1-8\exp \left(-\frac{t_j\Delta^2}{30\sigma^2}\right).$$
\end{lemma}

Using Lemma \ref{lemma_ub_phase_j}, we upper bound the probability that we make the incorrect decision in more than $1/4$ of the phases.

\begin{lemma}\label{lemma_prob_num_errors}
    Let $\eta < 1/4$ and set $C_1 = \frac{1}{600}, \,C_2 = 34$, then the following inequality holds under Algorithm \ref{alg:backtracking}.
    $$\mathbb{P}\left(\sum_{j=1}^J \mathds{1}\{G_j^C\} \geq J/4\right)  \leq \exp\left(-C_1\frac{\Delta^2}{\sigma^2}T + C_2 \log(\frac{1}{2\eta})\right).$$
\end{lemma}

 Putting Lemma \ref{lemma_ub_num_errors} and Lemma \ref{lemma_prob_num_errors} together we get the required result. We also assume $T> 60 \log (1/2\eta)$ so that we can even run the algorithm (playing each action in each phase at least once).
 \end{proof}

\subsection{Proof of Lemma \ref{lemma_ub_num_errors}}
\begin{proof}
Except for the fact that we are working in a continuous action space and not a finite one, the structure of our backtracking policy is similar to that of \citet{problem_dependent_threshold}. So the proof of this lemma is almost identical to what is seen in \citet{problem_dependent_threshold}. 

In our case, the objective that we would like to satisfy is a sufficiently accurate estimate of the change point $|\hat{x}_T-x^*|<\eta$ with high probability. As long as we make the correct decision in $\log_2(1/2\eta)$ more phases than we make incorrect decisions, then we will have made enough correct decisions to cancel out our bad ones - and correctly zoomed in enough to achieve $|\hat{x}_T-x^*|<\eta$. The quantity $\log_2(1/2\eta)$ is important as we need to correctly zoom in enough times such that the final region in the action space has sufficiently small width, i.e., $a_3^{J+1}-a_1^{J+1}<2\eta$ and contains the true change point $x^* \in [a_1^{J+1},a_3^{J+1})$. This ensures that estimating the midpoint of this region guarantees $|\hat{x}_T-x^*|<\eta$.\\

Writing this explicitly, it is sufficient to satisfy equation \eqref{eqn_bigT_ub_proof1_1} to achieve $|\hat{x}_T-x^*|<\eta$.
\begin{equation}\label{eqn_bigT_ub_proof1_1}
    \sum_{j=1}^J \mathds{1}\{G_j\} - \sum_{j=1}^J \mathds{1}\{G_j^C\} > \log_2(1/2\eta)
\end{equation}
Hence to prove this lemma, we just need to demonstrate that $\sum_{j=1}^J \mathds{1}\{G_j^C\} < J/4$ implies equation \eqref{eqn_bigT_ub_proof1_1} and therefore $|\hat{x}_T-x^*|<\eta$. 

We first note that if we have $\sum_{j=1}^J \mathds{1}\{G_j^C\} < J/4$, then we also have that $\sum_{j=1}^J \mathds{1}\{G_j\} \geq 3J/4$. We can then plug this into the following equation.

\begin{align}
    \sum_{j=1}^J \mathds{1}\{G_j^C\} - \sum_{j=1}^J \mathds{1}\{G_j^C\} &> \frac{3J}{4} - \frac{J}{4} = \frac{J}{2}\nonumber \nonumber\\
    &= \frac{1}{2} \left\lceil 6\log \left(\frac{1}{2\eta}\right) \right\rceil \label{eqn_bigT_ub_proof1_2}\\
    &= \frac{1}{2} \left\lceil 6\frac{\log_2 (\frac{1}{2\eta})}{\log_2 (e)} \right\rceil \nonumber\\
    &\geq \frac{1}{2} \left( 6\frac{\log_2 (\frac{1}{2\eta})}{\log_2 (e)} -1\right) \nonumber\\
    &>\log_2(1/2\eta)\nonumber
\end{align}
Where equation \eqref{eqn_bigT_ub_proof1_2} comes from the definition of $J = \left\lceil 6\log \left(\frac{1}{2\eta}\right) \right\rceil $ and the final inequality comes from the assumption that $\eta<1/4$.
\end{proof}

\newpage
\subsection{Proof of Lemma \ref{lemma_ub_phase_j}}
\begin{proof}
For this proof, we will additionally denote $\mathbb{P}_j(\cdot) = \mathbb{P}(\cdot|\mathcal{F}_{j-1})$.

From the definition of $G_j$ in equation \eqref{eq_Gj_def}, we have that 
\begin{align*}
    \mathbb{P}_j(G_j) \geq \min \bigg\{& \mathbb{P}_j\left(E_{P,j} \cap x^* \in (0,1) \backslash [a_1^j,a_3^j)\right),\\
    & \mathbb{P}_j\left(E_{P,j}^C \cap E_{L,j}^C \cap x^* \in [a_2^j,a_3^j)\right), \\
    &  \mathbb{P}_j\left(E_{P,j}^C \cap E_{R,j}^C \cap x^* \in  [a_1^j,a_2^j)\right) \bigg\}
\end{align*}

Now, since the events $\{x^* \in (0,1) \backslash [a_1^j,a_3^j)\},\, \{x^* \in [a_2^j,a_3^j)\}$ and $\{x^* \in  [a_1^j,a_2^j)\}$ are determined by the actions and rewards from the first $j-1$ phases in running the policy we have

\begin{align*}
    \mathbb{P}_j(G_j) \geq \min \{& \mathbb{P}_j\left(E_{P,j} \, \big|\, x^* \in (0,1) \backslash [a_1^j,a_3^j) \right) \mathds{1}\{ x^* \in (0,1) \backslash [a_1^j,a_3^j)\},\\
    & \mathbb{P}_j\left(E_{P,j}^C \cap E_{L,j}^C \, \big|\, x^* \in [a_2^j,a_3^j)\right) \mathds{1}\{x^* \in [a_2^j,a_3^j)\}, \\
    &  \mathbb{P}_j\left(E_{P,j}^C \cap E_{R,j}^C \, \big|\, x^* \in  [a_1^j,a_2^j)\right) \mathds{1}\{x^* \in  [a_1^j,a_2^j)\} \}\\
    \geq \min \{& \mathbb{P}_j\left(E_{P,j} \, \big|\, x^* \in (0,1) \backslash [a_1^j,a_3^j)\right),\\
     & \mathbb{P}_j\left(E_{P,j}^C \cap E_{L,j}^C \, \big|\, x^* \in [a_2^j,a_3^j)\right), \\
     &  \mathbb{P}_j\left(E_{P,j}^C \cap E_{R,j}^C \, \big|\, x^* \in  [a_1^j,a_2^j)\right) \}
\end{align*}

We can now use the following three lemmas to complete our proof.

\begin{lemma} \label{lemma_ub_phase_j_1}
    If $x^* \in (0,1) \backslash [a_1^j,a_3^j)$, then 
    $$\mathbb{P}_j\left(E_{P,j} \right) \geq 1 - 3 \exp \left(-\frac{t_j \Delta^2}{30\sigma^2}\right)$$
\end{lemma}
\begin{proof}

    Let's first assume that $x^* \in [a_3^j,1]$ and $\mu_1<\mu_2$. Now, if we have events $H,H_1,H_2,H_3$ such that $H_1\cap H_2 \cap H_3 \subset H$, then $\mathbb{P}(H^C) \leq \mathbb{P}(\{H_1\cap H_2 \cap H_3\}^C) = \mathbb{P}(H_1^C\cup H_2^C \cup H_3^C) \leq \mathbb{P}(H^C_1) + \mathbb{P}(H^C_2) + \mathbb{P}(H^C_3)$. We can then apply this as follows for equation \eqref{eq_ub_sets_implication}. But firstly, by definition of $E_{P,j}$ we can write the following.
    \begin{align}
        \mathbb{P}_j(E_{P,j}^C)  =& \mathbb{P}_j \Bigg(\left|\frac{\hat{\mu}_{0,t_j}+ \hat{\mu}_{a_1^j,t_j}}{2} - \frac{\hat{\mu}_{a_3^j,t_j}+\hat{\mu}_{1,t_j}}{2}\right| \\
        &\quad\qquad>\frac{3}{4}\max \left( \left| \frac{\hat{\mu}_{0,t_j} + \hat{\mu}_{a_1^j,t_j} + \hat{\mu}_{a_3^j,t_j}}{3} - \hat{\mu}_{1,t_j}\right| ,\, \left| \hat{\mu}_{0,t_j} - \frac{\hat{\mu}_{a_1^j,t_j} + \hat{\mu}_{a_3^j,t_j} + \hat{\mu}_{1,t_j}}{3}\right|     \right) \Bigg) \nonumber\\
        \leq & \mathbb{P}_j \left(\left|\frac{\hat{\mu}_{0,t_j}+ \hat{\mu}_{a_1^j,t_j}}{2} - \frac{\hat{\mu}_{a_3^j,t_j}+\hat{\mu}_{1,t_j}}{2}\right| > \frac{3}{4} \left| \frac{\hat{\mu}_{0,t_j} + \hat{\mu}_{a_1^j,t_j} + \hat{\mu}_{a_3^j,t_j}}{3} - \hat{\mu}_{1,t_j}\right|\right) \nonumber\\
        \leq & \mathbb{P}_j \left(\frac{\hat{\mu}_{0,t_j}+ \hat{\mu}_{a_1^j,t_j}}{2} > \frac{\hat{\mu}_{a_3^j,t_j}+\hat{\mu}_{1,t_j}}{2}\right) \nonumber\\
        & + \mathbb{P}_j \left(\frac{\hat{\mu}_{0,t_j} + \hat{\mu}_{a_1^j,t_j} + \hat{\mu}_{a_3^j,t_j}}{3} > \hat{\mu}_{1,t_j}\right) \label{eq_ub_sets_implication}\\
        &+ \mathbb{P}_j \left(-\frac{\hat{\mu}_{0,t_j}+ \hat{\mu}_{a_1^j,t_j}}{2} + \frac{\hat{\mu}_{a_3^j,t_j}+\hat{\mu}_{1,t_j}}{2} > \frac{3}{4} \left( -\frac{\hat{\mu}_{0,t_j} + \hat{\mu}_{a_1^j,t_j} + \hat{\mu}_{a_3^j,t_j}}{3} + \hat{\mu}_{1,t_j}\right)\right)  \nonumber\\
        \leq & \exp \left(-\frac{t_j \Delta^2}{8\sigma^2}\right) + \exp \left(-\frac{t_j \Delta^2}{24\sigma^2}\right) + \exp \left(-\frac{t_j \Delta^2}{30\sigma^2}\right) \label{eqn_bigT_ub_proof1_3}\\
        \leq & 3 \exp \left(-\frac{t_j \Delta^2}{30\sigma^2}\right) \label{eqn_bigT_ub_proof1_5}
    \end{align}
    As required. Where the penultimate inequality \eqref{eqn_bigT_ub_proof1_3} comes from upper bounding each of the three probabilities in equation \eqref{eq_ub_sets_implication}. To show these bounds hold, we will bound the first probability by rewriting it as shown below.
    \begin{align}
        \mathbb{P}_j \left(\frac{\hat{\mu}_{0,t_j}+ \hat{\mu}_{a_1^j,t_j}}{2} > \frac{\hat{\mu}_{a_3^j,t_j}+\hat{\mu}_{1,t_j}}{2}\right) &= \mathbb{P}_j \left(\hat{\mu}_{0,t_j}+ \hat{\mu}_{a_1^j,t_j} - \hat{\mu}_{a_3^j,t_j}+\hat{\mu}_{1,t_j} < 0\right) \nonumber\\
        &= \mathbb{P}_j \left(\hat{\mu}_{0,t_j}+ \hat{\mu}_{a_1^j,t_j} - \hat{\mu}_{a_3^j,t_j}+\hat{\mu}_{1,t_j} - \Delta < -\Delta \right) \label{eqn_bigT_ub_proof1_4}\\ 
        &\leq \exp \left(-\frac{t_j \Delta^2}{8\sigma^2}\right)\nonumber
    \end{align}
Where the final inequality comes from noticing that, conditioning on the action rewards from the first $j-1$ phases $\mathcal{F}_{j-1}$, the empirical means $\hat{\mu}_{0,t_j}, \hat{\mu}_{a_1^j,t_j}, \hat{\mu}_{a_3^j,t_j},\hat{\mu}_{1,t_j}$ are independent, each $\sigma^2/t_j$-sub-Gaussian, and with respective means $\mu_1,\mu_1,\mu_1,\mu_2$. Hence, the sum $\hat{\mu}_{0,t_j}+ \hat{\mu}_{a_1^j,t_j} - \hat{\mu}_{a_3^j,t_j}+\hat{\mu}_{1,t_j} - \Delta$ is mean-zero, $4\sigma^2/t_j$-sub-Gaussian. Plugging this into the Hoeffding inequality from Proposition 2.5 in \cite{wainwright_2019}, we attain the final inequality above.

For the final two probabilities in equation \eqref{eq_ub_sets_implication} we can similarly calculate the upper bounds as following to attain equation \eqref{eqn_bigT_ub_proof1_3}.
    \begin{align*}
     \mathbb{P}_j \left(\frac{\hat{\mu}_{0,t_j} + \hat{\mu}_{a_1^j,t_j} + \hat{\mu}_{a_3^j,t_j}}{3} > \hat{\mu}_{1,t_j}\right)&\leq \exp \left(-\frac{t_j \Delta^2}{24\sigma^2}\right)\\
    \mathbb{P}_j \left(-\frac{\hat{\mu}_{0,t_j}+ \hat{\mu}_{a_1^j,t_j}}{2} + \frac{\hat{\mu}_{a_3^j,t_j}+\hat{\mu}_{1,t_j}}{2} > \frac{3}{4} \left( -\frac{\hat{\mu}_{0,t_j} + \hat{\mu}_{a_1^j,t_j} + \hat{\mu}_{a_3^j,t_j}}{3} + \hat{\mu}_{1,t_j}\right)\right) &\leq \exp \left(-\frac{t_j \Delta^2}{30\sigma^2}\right)
    \end{align*}

Hence, returning to equation \eqref{eqn_bigT_ub_proof1_5} we can achieve the following final bound as required for the lemma.
\begin{equation} \label{eqn_bigT_ub_proof1_6}
    \Longrightarrow  \mathbb{P}_j(E_{P,j}) >  1 - 3 \exp \left(-\frac{t_j \Delta^2}{30\sigma^2}\right)
\end{equation}

If we instead assume that $x^* \in [0,a_1^j)$ then almost identical arguments would lead us to the same final equation \eqref{eqn_bigT_ub_proof1_6}. Furthermore if we were to assume that $\mu_1>\mu_2$ then we could again use almost identical arguments in both cases to attain  \eqref{eqn_bigT_ub_proof1_6}.

\end{proof}

\begin{lemma}\label{lemma_ub_phase_j_2}
    If $x^* \in [a_2^j,a_3^j)$, then
    $$\mathbb{P}_j\left(E_{P,j}^C \cap E_{L,j}^C\right) \geq 1- 8 \exp \left(-\frac{t_j \Delta^2}{30\sigma^2}\right)$$
\end{lemma}
\begin{proof}

Note that by a union bound we have
\begin{equation}\label{eqn_bigT_ub_proof1_10}
    \mathbb{P}_j(\{E_{P,j}^C\cap E_{L,j}^C\}^C)  = \mathbb{P}_j(E_{P,j}\cup E_{L,j}) \leq \mathbb{P}_j(E_{P,j}) +\mathbb{P}_j(E_{L,j})
\end{equation}

So we will focus on bounding the two probabilities on the right of the above equation. Again, we will first assume that $\mu_1<\mu_2$.\\ 

\textbf{Starting with} $\mathbb{P}_j(E_{P,j})$, we can use a union bound to get the following.

\begin{align}
\mathbb{P}_j(E_{P,j})   =& \mathbb{P}_j \Bigg(\left|\frac{\hat{\mu}_{0,t_j}+ \hat{\mu}_{a_1^j,t_j}}{2} - \frac{\hat{\mu}_{a_3^j,t_j}+\hat{\mu}_{1,t_j}}{2}\right|\\ 
& \quad \qquad<\frac{3}{4}\max \left( \left| \frac{\hat{\mu}_{0,t_j} + \hat{\mu}_{a_1^j,t_j} + \hat{\mu}_{a_3^j,t_j}}{3} - \hat{\mu}_{1,t_j}\right| ,\, \left| \hat{\mu}_{0,t_j} - \frac{\hat{\mu}_{a_1^j,t_j} + \hat{\mu}_{a_3^j,t_j} + \hat{\mu}_{1,t_j}}{3}\right|     \right) \Bigg) \nonumber\\
\begin{split}\label{eqn_bigT_ub_proof1_7}
    \leq &\mathbb{P}_j\left(\left|\frac{\hat{\mu}_{0,t_j}+ \hat{\mu}_{a_1^j,t_j}}{2} - \frac{\hat{\mu}_{a_3^j,t_j}+\hat{\mu}_{1,t_j}}{2}\right| < \frac{3}{4} \left| \frac{\hat{\mu}_{0,t_j} + \hat{\mu}_{a_1^j,t_j} + \hat{\mu}_{a_3^j,t_j}}{3} - \hat{\mu}_{1,t_j}\right|   \right)\\
& + \mathbb{P}_j\left(\left|\frac{\hat{\mu}_{0,t_j}+ \hat{\mu}_{a_1^j,t_j}}{2} - \frac{\hat{\mu}_{a_3^j,t_j}+\hat{\mu}_{1,t_j}}{2}\right| < \frac{3}{4} \left| \hat{\mu}_{0,t_j} - \frac{\hat{\mu}_{a_1^j,t_j} + \hat{\mu}_{a_3^j,t_j} + \hat{\mu}_{1,t_j}}{3}\right|    \right)
\end{split}
\end{align}

We can then denote the final two probabilities in equation \eqref{eqn_bigT_ub_proof1_7} as $(A')$ and $(B')$ respectively. We can firstly bound $(A')$ using a similar idea to equation \eqref{eq_ub_sets_implication}. Namely, if we have events $H,H_1,H_2,H_3$ such that $H_1\cap H_2 \cap H_3 \subset H$, then $\mathbb{P}(H^C) \leq \mathbb{P}(\{H_1\cap H_2 \cap H_3\}^C) = \mathbb{P}(H_1^C\cup H_2^C \cup H_3^C) \leq \mathbb{P}(H^C_1) + \mathbb{P}(H^C_2) + \mathbb{P}(H^C_3)$. We use this to get the below equation.

\begin{align}
    (A') \leq & \mathbb{P}_j\left(\frac{\hat{\mu}_{0,t_j}+ \hat{\mu}_{a_1^j,t_j}}{2} >\frac{\hat{\mu}_{a_3^j,t_j}+\hat{\mu}_{1,t_j}}{2}\right)\nonumber\\
     & + \mathbb{P}_j\left(  \frac{\hat{\mu}_{0,t_j} + \hat{\mu}_{a_1^j,t_j} + \hat{\mu}_{a_3^j,t_j}}{3} > \hat{\mu}_{1,t_j}   \right)\label{eqn_bigT_ub_proof1_8}\\
     & + \mathbb{P}_j\left( -\frac{\hat{\mu}_{0,t_j}+ \hat{\mu}_{a_1^j,t_j}}{2} + \frac{\hat{\mu}_{a_3^j,t_j}+\hat{\mu}_{1,t_j}}{2} < \frac{3}{4} \left(-\frac{\hat{\mu}_{0,t_j} + \hat{\mu}_{a_1^j,t_j} + \hat{\mu}_{a_3^j,t_j}}{3} + \hat{\mu}_{1,t_j}\right)\right)\nonumber\\
     \leq & 3\exp \left(-\frac{t_j \Delta^2}{24\sigma^2}\right) \nonumber
\end{align}
Where the final line again comes from using the Hoeffding inequality for each of the probabilities in \eqref{eqn_bigT_ub_proof1_8}, similar to the previous lemma for equation \eqref{eqn_bigT_ub_proof1_4}.

Furthermore, we can similarly show that $(B') \leq 3\exp \left(-\frac{t_j \Delta^2}{24\sigma^2}\right)$. Hence, plugging our bounds for $(A')$ and $(B')$ into equation \eqref{eqn_bigT_ub_proof1_7} we have the following.
\begin{equation}\label{eqn_bigT_ub_proof1_9}
    \Longrightarrow \mathbb{P}_j(E_{P,j}) \leq 6 \exp \left(-\frac{t_j \Delta^2}{24\sigma^2}\right)
\end{equation}\\

\textbf{Now let's consider} $\mathbb{P}_j(E_{L,j})$. By noting that $\{\hat{\mu}_{a_1^j,t_j}<\hat{\mu}_{a_3^j,t_j}\} \cap \{\hat{\mu}_{a_2^j,t_j} - \hat{\mu}_{a_1^j,t_j} < \hat{\mu}_{a_3^j,t_j} - \hat{\mu}_{a_2^j,t_j}\} \Longrightarrow \{|\hat{\mu}_{a_1^j,t_j} - \hat{\mu}_{a_2^j,t_j}| < |\hat{\mu}_{a_2^j,t_j} - \hat{\mu}_{a_3^j,t_j}|\}$, we can bound the following probability. 
    \begin{align}
    \mathbb{P}_j(E_{L,j}) &=
        \mathbb{P}( \{|\hat{\mu}_{a_1^j,t_j} - \hat{\mu}_{a_2^j,t_j}| < |\hat{\mu}_{a_2^j,t_j} - \hat{\mu}_{a_3^j,t_j}|\}^C) \nonumber\\
        & \leq \mathbb{P}\left(\left\{\{\hat{\mu}_{a_1^j,t_j}<\hat{\mu}_{a_3^j,t_j}\} \cap \{\hat{\mu}_{a_2^j,t_j} - \hat{\mu}_{a_1^j,t_j} < \hat{\mu}_{a_3^j,t_j} - \hat{\mu}_{a_2^j,t_j}\}\right\}^C\right)\nonumber\\
        & \leq \mathbb{P}(\{\hat{\mu}_{a_1^j,t_j}<\hat{\mu}_{a_3^j,t_j}\}^C) + \mathbb{P}(\{\hat{\mu}_{a_2^j,t_j} - \hat{\mu}_{a_1^j,t_j} < \hat{\mu}_{a_3^j,t_j} - \hat{\mu}_{a_2^j,t_j}\}^C)\nonumber\\
        &\leq \exp\left(-\frac{t_j \Delta^2}{4\sigma^2}\right) + \exp\left(-\frac{t_j \Delta^2}{12\sigma^2}\right)\nonumber\\
        & \leq 2 \exp \left(-\frac{t_j \Delta^2}{12\sigma^2}\right)\label{eqn_used_for_smallT_ub}
    \end{align}
    Where the penultimate inequality again holds using Hoeffding inequality as in equation \eqref{eqn_bigT_ub_proof1_4}. Hence, plugging this and equation \eqref{eqn_bigT_ub_proof1_9} into \eqref{eqn_bigT_ub_proof1_10} give us the following.

    \begin{align*}
        \mathbb{P}_j(\{E_{P,j}^C\cap E_{L,j}^C\}^C)  & \leq \mathbb{P}_j(E_{P,j}) +\mathbb{P}_j(E_{L,j})\\
        &\leq  8 \exp \left(-\frac{t_j \Delta^2}{30\sigma^2}\right)
    \end{align*}
As required. We note that one could use an almost identical method to attain the same bound if $\mu_1>\mu_2$.

\end{proof}

\begin{lemma} \label{lemma_ub_phase_j_3}
    If $x^* \in  [a_1^j,a_2^j)$, then
    $$ \mathbb{P}_j\left(E_{P,j}^C \cap E_{R,j}^C\right) \geq 1- 8 \exp \left(-\frac{t_j \Delta^2}{30\sigma^2}\right)$$
\end{lemma}
\begin{proof}
    Similar to Lemma \ref{lemma_ub_phase_j_2}
\end{proof}

    Hence, regardless of the position of the change point, we get the required bound for the probability of zooming in/out towards the correct direction, given the actions and rewards in previous phases, concluding our proof of Lemma \ref{lemma_ub_phase_j}.
    
\end{proof}
\subsection{Proof of Lemma \ref{lemma_prob_num_errors}}
\begin{proof}
Again, the structure of this proof will be very similar to the proofs seen in \citet{problem_dependent_threshold} since the structure of the policies is very similar. Of course, the probability of a good event in each round will be different as well as constants throughout.
    \begin{align*}
        p_j &:= \mathbb{P}_j (G_j^C)\\
        p_0 &:= 8\exp \left(-\frac{t_j\Delta^2}{30\sigma^2}\right)
    \end{align*}

We can furthermore assume that
\begin{equation}\label{eq_Delta_assumption}
    \Delta \geq 88\sqrt{\frac{\sigma^2\log(1/2\eta)}{T}}
\end{equation}

Otherwise, if we were to assume the contrary, then plugging this into the exponential term in Theorem \ref{fixed budget upper bound backtracking}, the result from Theorem \ref{fixed budget upper bound backtracking} would trivially be true. Now, using Lemma \ref{lemma_ub_phase_j} and substituting this in our definition of $p_0$ we have
\begin{align}
    p_j \leq p_0 &= 8\exp \left(-\frac{t_j\Delta^2}{30\sigma^2}\right) \nonumber\\
    &\leq 8\exp \left(-88^2 \log (1/2\eta) \frac{t_j}{30T}\right)\label{eqn_bigT_ub_proof1_11}\\
    &\leq 8\exp \left(-88^2 \log (1/2\eta)  \frac{T}{10J} \frac{1}{30T}\right)\label{eqn_bigT_ub_proof1_12}\\
    &= 8\exp \left(-88^2 \log (1/2\eta)   \frac{1}{300J}\right)\nonumber\\
    &\leq 8\exp \left(-88^2 \log (1/2\eta)   \frac{1}{300(6\log (1/2\eta) + 1)}\right)\label{eqn_bigT_ub_proof1_13}\\
    &\leq \frac{1}{4} \label{eqn_bigT_ub_proof1_18}
\end{align}
Where equation \eqref{eqn_bigT_ub_proof1_11} come from substituting assumption on $\Delta$ in \eqref{eq_Delta_assumption}. Equation \eqref{eqn_bigT_ub_proof1_12} comes from using the definition of $t_j=\lfloor \frac{T}{5J}\rfloor \geq \frac{T}{10J}$ (since we assume that $\frac{T}{5J} \geq 1$ for the policy to have enough samples to run). Equation \eqref{eqn_bigT_ub_proof1_13} comes from the definition of $J = \lceil 6\log  (1/2\eta) \rceil \leq 6\log  (1/2\eta) + 1$. Then, the final inequality holds true whenever  $\eta < 1/4$.

Now, the quantity of interest in this lemma is the probability that we fail in more than $1/4$ of the phases. We can bound this probability from above using Markov's inequality and any $\lambda \geq 0$.

\begin{equation}\label{eq_ub_markov_bound}
    \mathbb{P}\left(\sum_{j=1}^J \mathds{1}\{G_j^C\} \geq J/4\right) \leq \frac{\mathbb{E} \left[\exp \left(\lambda \sum_{j=1}^J \mathds{1}\{G_j^C\} \right)\right]}{\exp (\lambda \cdot \frac{J}{4})}
\end{equation}
Now, in order to bound the expectation on the right hand side of the above equation as in \citet{problem_dependent_threshold}, we will introduce the following function $\phi_p(\lambda) = \log(1-p+pe^\lambda)$ which is non-decreasing in $p$. Hence, since $p_j \leq p_0$ from Lemma \ref{lemma_ub_phase_j}, we have $\phi_{p_t}(\lambda) \leq \phi_{p_0}(\lambda)$. Hence we have the following, starting by using the tower rule.
\begin{align}
    \mathbb{E} \left[\exp \left(\lambda \sum_{j=1}^J \mathds{1}\{G_j^C\} \right)\right] & = \mathbb{E} \left[ \mathbb{E}_J\left[\exp \left(\lambda \mathds{1}\{G_J^C\}\right)\right]    \exp \left(\lambda \sum_{j=1}^{J-1} \mathds{1}\{G_j^C\} \right)\right] \nonumber\\
    &=\mathbb{E} \left[ \exp \left(\phi_{p_J}(\lambda)\right) \exp \left(\lambda \sum_{j=1}^{J-1} \mathds{1}\{G_j^C\} \right)\right]\label{eqn_bigT_ub_proof1_14}\\
    &\leq  \mathbb{E} \left[  \exp \left(\phi_{p_0}(\lambda)\right) \exp \left(\lambda \sum_{j=1}^{J-1} \mathds{1}\{G_j^C\} \right)\right]\label{eqn_bigT_ub_proof1_15}\\
    & \leq \mathbb{E} \left[  \exp \left(J \cdot \phi_{p_0}(\lambda)\right)\right]\nonumber
\end{align}
Where equation \eqref{eqn_bigT_ub_proof1_14} comes from the definition $p_J= \mathbb{P}_J (G_J^C)$ and hence that $\mathbb{E}_J\left[\exp \left(\lambda \mathds{1}\{G_J^C\}\right)\right] = \exp(\lambda)p_J + 1-p_J = \exp \left(\phi_{p_J}(\lambda)\right)$. Equation \eqref{eqn_bigT_ub_proof1_15} comes from the monotonicity of $\phi$ and we can get the final inequality by simply repeating these steps.

We can then substitute this bound in expectation back into the inequality from \eqref{eq_ub_markov_bound}, remembering that we can choose any $\lambda \geq 0$

\begin{align}
    \mathbb{P}\left(\sum_{j=1}^J \mathds{1}\{G_j^C\}\right) &\leq \exp \left(-J \cdot \sup_{\lambda \geq 0} \left(\lambda/4 - \phi_{p_0}(\lambda)\right)\right) \nonumber\\
    &= \exp \left(-J \cdot kl(1/4,p_0)\right) \label{eqn_bigT_ub_proof1_16}\\
    &\leq \exp \left(-J\frac{1}{4}\log (\frac{1}{p_0}) + J\log(2)\right)\label{eqn_bigT_ub_proof1_17}\\
    &\leq \exp\left( - \frac{Jt_j}{120} \frac{\Delta^2}{\sigma^2} + J\left(\log(2)+\frac{\log(8)}{4}\right) \right)\nonumber\\
    &\leq \exp\left(   -\frac{T\Delta^2}{600\sigma^2} + 13\log(1/2\eta) \right)\nonumber
\end{align}
As required. Where equation \eqref{eqn_bigT_ub_proof1_16} comes from noting that $\sup_{\lambda \geq 0} \lambda q - \phi_p(\lambda) =kl(q,p)$ when $q \geq p$, where we denote $kl(p,q)$ as the divergence between two Bernoulli distributions of parameters $p$ and $q$. We can then use this and the fact that $p_0 \leq 1/4$ from \eqref{eqn_bigT_ub_proof1_18} to attain equation \eqref{eqn_bigT_ub_proof1_16}. Then we can attain equation \eqref{eqn_bigT_ub_proof1_17} from the trick $kl(a,b) \geq a\log(1/b)-\log(2)$.

\end{proof}

\newpage
\section{Proofs For Large Budget Lower Bound}\label{app_largeT_lowerbound_proofs}

\subsection{Overview}

Before going into technical details, we provide a brief informal overview of the techniques we use to prove the lower bound in Theorem \ref{corr_large_budget_lb}. In particular, we also discuss how $\eta$ appears through our analysis, unlike previous works (see Section \ref{section_largeT_lower_bound} for discussion). We first simplify the problem and assume that the two means ($\mu_1,\mu_2$) are known and that the change point is at one of $K=\lfloor1/2\eta\rfloor$ different positions, $\lbrace x_j^* \rbrace_j$ indexed from smallest to largest, each at least $2\eta$ away from each other. Thus our objective is to identify which of these $K$ prospective environments $V’ = \lbrace v_ j \rbrace_ j$ we are in. The proof then uses the following three ingredients.

 \begin{enumerate}[(i)]
     \item \textbf{Conditions on the policy:} The piecewise constant structure makes it difficult to isolate the effect of the expected number of plays in each region of $\mathcal{A}$ with respect to the failure probability (whereas this is possible for each arm in Best Arm Identification , see \citet{Carpentier2016TightLowerBounds} equation (7), and in unstructured Thresholding Bandits, see \citet{locatelli16_thresholding} appendix A.1). Indeed since the two means are known, in order to distinguish $v_i$ from all other environments in $V’$, it is sufficient to sample exclusively on either side adjacent to the change point of $v_i$, $x^*_i$. However, this is infeasible in practice as the learner will not a priori know where the changepoint $x^*_i$ occurs. To enforce some exploration across the action space, we begin by making a reasonable assumption that the failure probability of our policy is at most $C’=1/2$ regardless of the environment (See Theorem \ref{theorem_lb} statement). Note in Appendix \ref{appendix_section_large_budgets_lower_bound_finishing_proof} we show that we can omit this condition when the budget is sufficiently large.
     \item \textbf{Flat environment $v_0$:} 
     We consider a flat environment $v_0$ with no change in mean. Here there will always be a region in which $\pi$ estimates the change point with probability less than $1/(K-1)$ (Lemma \ref{lemma_lb_exists_i}). We then choose a challenging reference environment $v_i \in V'$ to have a change point within this low probability region.
     \item \textbf{Change of measure and transportation lemma:} Using a change of measure from $v_K$ to $v_i$, we can relate the number of samples played to the right of the change point in environment $v_i$, denoted $T_{(x_i^*,1]}$, with the failure probability (see equations \eqref{eqn_bigT_lb_proof4} - \eqref{eqn_bigT_lb_proof7}). For a policy to be able to identify a change in mean with high probability, we require a sufficient number of samples to the right \emph{and} left of the change (missing this is why some existing policies can suffer when the change is near the boundary, see Section \ref{section_experiments}). Therefore, using (i), we can upper bound $T_{(x_i^*,1]}$. This is firstly formalised with a change of measure from $v_i$ to $v_0$. Then we use a transportation lemma (see Lemma \ref{lemma_lb_KL_2} proof) combined with the $1/(K-1)$ probability in (ii) and $C’$ from (i) (see Lemma \ref{lemma_lb_KL_2} and equations \eqref{eqn_bigT_lb_proof8}-\eqref{eqn_bigT_lb_proof9}) to produce our lower bound. Note that the definition of $K$ here involves $\eta$.
 \end{enumerate}

\subsection{Lower Bound for Reasonable Policies}
In line with what is outlined above, we begin by considering a lower bound on any policy which explores sufficiently. In Appendix~\ref{appendix_section_large_budgets_lower_bound_finishing_proof} we will show how this assumption can be removed to get our final lower bound. 
\begin{theorem}\label{theorem_lb}
 Let $\Bar{V} \subset V(\Delta,\sigma)$ be the set of environments with change in mean $\Delta$ and Gaussian random noise with variance $\sigma^2$. Let $C' \in (0,1)$ and
     let $\Bar{\Pi} := \{\pi \in \Pi : \forall v \in \Bar{V},\, \mathbb{P}_{v,\pi} (|\hat{x}_T - x^*_v| < \eta) \geq C'\}  \subset \Pi$. Then, denoting $x^*_v$ as the change point in environment $v$, we have
    \begin{align*}
        \inf_{\pi \in \Bar{\Pi}} \sup_{v \in \Bar{V}} \mathbb{P}_{v,\pi} (|\hat{x}_T - x^*_v| \geq \eta)
        \geq \frac{1}{8}\exp \left(-\frac{\Delta^2}{2\sigma^2}T + C' \log \left(\left\lfloor\frac{1}{2\eta}\right\rfloor-1\right)\right).
    \end{align*}
\end{theorem}
\begin{proof}
Fix some $\pi \in \Pi$. Let $\Bar{V}' := \{v_1,...,v_K\}$ with $K := \lfloor1/2\eta\rfloor$ be a finite subset of $\Bar{V}$ in which all environments have mean rewards $\mu_1,\mu_2$ and noise variance $\sigma^2$. Furthermore, let environment $v_j$ have change point $x^*_j := 2\eta(j-1)$. Finally we will drop the subscript for $\pi$ as we will fix this policy $\pi$ for the remained or the proof, unless otherwise stated (i.e. denote $\mathbb{P}_{v_j}=\mathbb{P}_{v_j,\pi}$).  Then we have
\begin{equation}\label{eq_lb_restricting_sup1}
    \sup_{v \in \Bar{V}} \mathbb{P}_{v,\pi} (|\hat{x}_T - x^*_v| \geq \eta) \geq \sup_{v_j \in \Bar{V}'} \mathbb{P}_{v_j} (|\hat{x}_T - x^*_j| \geq \eta)
\end{equation}

Now, let $v_0$ be an environment in which there is no change in mean across the space. Namely the mean reward function has constant value $\mu_2$. We will now introduce several helpful lemmas before continuing with equation \eqref{eq_lb_restricting_sup1}. 

\begin{lemma}\label{lemma_lb_exists_i}
    There exists an $i \in \{1,...,K-1\}$ such that
    $$\mathbb{P}_{v_0} (|\hat{x}_T - x^*_i| < \eta) \leq \frac{1}{K-1}$$
\end{lemma}
\begin{proof}
    Suppose, for contradiction, that $\forall i \in \{1,..,K-1\}$
    \begin{align}
        \mathbb{P}_{v_0} (|\hat{x}_T - x^*_i| < \eta) &> \frac{1}{K-1}\\
         \Longrightarrow 1 \geq \sum_{i=1}^{K-1}\mathbb{P}_{v_0} (|\hat{x}_T - x^*_i| < \eta) &> \sum_{i=1}^{K-1}\frac{1}{K-1}=1
    \end{align}
    Which is a contradiction, hence the proof is complete. Note the final inequality on the left holds because the sets $\{(x^*_i-\eta,x^*_i+\eta)\}_{i=1}^{K-1}$ are disjoint and  $\mathbb{P}_{v_0}$ is a probability measure.
\end{proof}

Hence, we set $i$ to be the same $i$ as in the above Lemma for the remainder of the proof.\\

\begin{lemma}\label{lemma_lb_cont_KL}
    Let $T_B$ be the number of times we play an action in the set $B \subseteq [0,1]$ over the whole budget $T$. Then, with $i$ chosen to satisfy Lemma \ref{lemma_lb_exists_i},
    $$KL(\mathbb{P}_{v_i},\mathbb{P}_{v_0}) \leq \frac{\Delta^2}{2\sigma^2}\mathbb{E}_{v_i}\left[T_{[0,x^*_i]}\right].$$
\end{lemma}
\begin{proof}
    Denote $P_{i,A_t}$ and $P_{0,A_t}'$ as the reward distributions of the action played in round $t$ in both the environment $v_i$ and $v_0$, respectively. Then, from Ex 15.8 \citep{BanditAlgosBook}, we have the following equation.
    \begin{align}
        KL(\mathbb{P}_{v_i},\mathbb{P}_{v_0}) &= \mathbb{E}_{v_i}\left[\sum_{t=1}^T KL(P_{i,A_t}, P_{0,A_t} )\right]\nonumber\\
        &= \mathbb{E}_{v_i}\left[\sum_{t=1}^T KL(P_{i,A_t}, P_{0,A_t}) \cdot (\mathds{1}\{A_t \in [0,x^*_i]\} + \mathds{1}\{A_t \in (x^*_i,1]\})\right]\nonumber\\
        &= \mathbb{E}_{v_i}\left[\sum_{t=1}^T KL(P_{i,A_t}, P_{0,A_t}) \mathds{1}\{A_t \in [0,x^*_i]\}\right]\label{eqn_bigT_lb_proof1}\\
        &= \mathbb{E}_{v_i}\left[\sum_{t=1}^T \frac{\Delta^2}{2\sigma^2}\mathds{1}\{A_t \in [0,x^*_i]\}\right]\label{eqn_bigT_lb_proof2}\\
        &= \frac{\Delta^2}{2\sigma^2}\mathbb{E}_{v_i}\left[T_{[0,x^*_i]}\right]\nonumber
    \end{align}
    As required. Where equation \eqref{eqn_bigT_lb_proof1} comes from noting that for $A_t \in (x^*_i,1]$, we have $P_{i,A_t} = P_{0,A_t}$, hence $KL(P_{i,A_t}, P_{0,A_t})=0$. Furthermore \eqref{eqn_bigT_lb_proof2} comes from noting that for $A_t \in [0,x^*_i]$ we are comparing $P_{i,A_t}, P_{0,A_t}$ which are two Gaussian distributions with difference in mean $\Delta$ and same variance $\sigma^2$.
\end{proof}

\begin{lemma}\label{lemma_lb_cont_KL2}
    Under the same setup as Lemma \ref{lemma_lb_cont_KL},
    $$KL(\mathbb{P}_{v_i},\mathbb{P}_{v_K}) \leq \mathbb{E}_{v_i}[T_{(x^*_i,x^*_K]}] \frac{\Delta^2}{2\sigma^2}.$$
\end{lemma}
\begin{proof}
    Similar to Lemma \ref{lemma_lb_cont_KL}, noting instead that the reward distributions in environments $v_i,v_K$ only differ in the region in the action space $(x^*_i,x^*_K]$.
\end{proof}

 \begin{lemma} \label{lemma_lb_KL_2}
     $$KL(\mathbb{P}_{v_i},\mathbb{P}_{v_0}) \geq C' \log \left((K-1)\right) - \log (2)$$
 \end{lemma}
 \begin{proof}
 Firstly, we have from Lemma 1 in \cite{Garivier_chain_rule_trick}, that the following holds for any measurable function $Z$ which maps to $[0,1]$.

 \begin{equation*}
     KL(\mathbb{P}_{v_i},\mathbb{P}_{v_0}) \geq kl(\mathbb{E}_{v_i}(Z),\mathbb{E}_{v_0}(Z))
 \end{equation*}
 
Now, if we choose $Z = \mathds{1}\{E\}$ and let the event $E := \{|\hat{x}_T - x^*_i| < \eta\}$, then this becomes the below equation.

 \begin{align}
    KL(\mathbb{P}_{v_i},\mathbb{P}_{v_0}) &\geq kl(\mathbb{P}_{v_i}(E),\mathbb{P}_{v_0}(E)) \nonumber\\
    &\geq \mathbb{P}_{v_i}(E) \log \left(\frac{1}{\mathbb{P}_{v_0}(E)}\right) - \log (2) \label{eqn_bigT_lb_proof3}\\
    & \geq C' \log \left(K-1\right) - \log (2)
\end{align}
Where equation \eqref{eqn_bigT_lb_proof3} comes from using a trick to bound $kl(a,b) \geq a \log (\frac{1}{b})-\log(2)$.  The the final inequality holds by  Lemma \ref{lemma_lb_exists_i} that $\mathbb{P}_{v_0} (E) \leq \frac{1}{K-1}$ and using the assumption made for $\pi$ in Theorem \ref{theorem_lb} that $\forall v \in \Bar{V},\, \mathbb{P}_{v,\pi} (|\hat{x}_T - x^*_v| < \eta) \geq C'$.
 \end{proof}

 Now, we put Lemmas \ref{lemma_lb_exists_i}, \ref{lemma_lb_cont_KL}, \ref{lemma_lb_cont_KL2}, and \ref{lemma_lb_KL_2} together and return to equation \eqref{eq_lb_restricting_sup1}.
\begin{align}
    \sup_{v \in \Bar{V}} \mathbb{P}_{v,\pi} (|\hat{x}_T - x^*_v| \geq \eta) &\geq \sup_{v_j \in \Bar{V}'} \mathbb{P}_{v_j} (|\hat{x}_T - x^*_j| \geq \eta)\nonumber\\
    &\geq \frac{1}{2}\mathbb{P}_{v_i}(|\hat{x}_T - x^*_i| \geq \eta) + \frac{1}{2}\mathbb{P}_{v_{K}}(|\hat{x}_T - x^*_K| \geq \eta)\label{eqn_bigT_lb_proof4}\\
    &\geq \frac{1}{2}\mathbb{P}_{v_i}(|\hat{x}_T - x^*_i| \geq \eta) + \frac{1}{2}\mathbb{P}_{v_{K}}(|\hat{x}_T - x^*_i| < \eta)\label{eqn_bigT_lb_proof5_1}\\
    &\geq \frac{1}{4} \exp \left(-KL(\mathbb{P}_{v_i},\mathbb{P}_{v_{K}})\right)\label{eqn_bigT_lb_proof5_2}\\
    &\geq \frac{1}{4} \exp \left(- \mathbb{E}_{v_i}[T_{(x^*_i,x^*_K]}] \frac{\Delta^2}{2\sigma^2}\right) \label{eqn_bigT_lb_proof6}\\
    &\geq \frac{1}{4} \exp \left(-\frac{\Delta^2}{2\sigma^2}T + \frac{\Delta^2}{2\sigma^2}\left( \mathbb{E}_{v_i}[T_{[0,x^*_i]}]  + \mathbb{E}_{v_i}[T_{(x^*_K,1]}]  \right)\right) \label{eqn_bigT_lb_proof7}\\
    &\geq \frac{1}{4} \exp \left(-\frac{\Delta^2}{2\sigma^2}T + \frac{\Delta^2}{2\sigma^2}\mathbb{E}_{v_i}[T_{[0,x^*_i]}]\right)\label{eqn_bigT_lb_proof8}\\
    &\geq \frac{1}{4} \exp \left(-\frac{\Delta^2}{2\sigma^2}T + KL(\mathbb{P}_{v_i},\mathbb{P}_{v_0})\right)\nonumber\\
    &\geq \frac{1}{4} \exp \left(-\frac{\Delta^2}{2\sigma^2}T + C' \log \left(K-1\right) - \log (2)\right) \label{eqn_bigT_lb_proof9}
\end{align}
As required. Inequality \eqref{eqn_bigT_lb_proof4} holds as we are simply taking the average of two point in the set over which the supremum is acting. Inequality \eqref{eqn_bigT_lb_proof5_1} holds since $|x^*_K - x^*_i| \geq 2\eta$ and therefore $\{|\hat{x}_T - x^*_i| < \eta\}\subset\{|\hat{x}_T - x^*_K| \geq \eta\}$. Inequality \eqref{eqn_bigT_lb_proof5_2} comes from the Bretagnolle-Huber inequality \citep{BanditAlgosBook} Theorem 14.2. Inequality \eqref{eqn_bigT_lb_proof6} comes from Lemma \ref{lemma_lb_cont_KL2}. Equation \eqref{eqn_bigT_lb_proof7} is true since $T= T_{[0,x^*_i]}+ T_{(x^*_i,x^*_K]} + T_{(x^*_K,1]}$. Inequality \eqref{eqn_bigT_lb_proof8} holds since $T_{(x^*_K,1]} \geq 0$. Then the final two inequalities hold from Lemmas \ref{lemma_lb_cont_KL} and \ref{lemma_lb_KL_2}, respectively.  
\end{proof}

\subsection{Proof of Theorem \ref{corr_large_budget_lb}}\label{appendix_section_large_budgets_lower_bound_finishing_proof}
We can extend Theorem \ref{theorem_lb} to the set of \emph{any} policies in Theorem \ref{corr_large_budget_lb_original_threshold} for a sufficiently large budget $T \geq  \frac{\sigma^2}{\Delta^2}  \log \left(\frac{1}{2}(\lfloor1/2\eta\rfloor-1)\right)$, which is stated below.
Theorem \ref{corr_large_budget_lb} can be seen as a consequence of Theorem \ref{corr_large_budget_lb_original_threshold} below since we have that $$ \frac{\sigma^2}{\Delta^2}  \log \left(\frac{1}{2}(\lfloor1/2\eta\rfloor-1)\right) \leq \frac{\sigma^2}{\Delta^2}(1.59\log(\lfloor\frac{1}{2\eta}\rfloor)-2\log(2)).$$
Therefore, since Theorem \ref{corr_large_budget_lb_original_threshold} holds for $T \geq  \frac{\sigma^2}{\Delta^2}  \log \left(\frac{1}{2}(\lfloor1/2\eta\rfloor-1)\right)$, it will also hold for $T \geq  \frac{\sigma^2}{\Delta^2}(1.59\log(\lfloor\frac{1}{2\eta}\rfloor)-2\log(2))$, as required.\\

\begin{theorem}\label{corr_large_budget_lb_original_threshold}
     Let $\Bar{V} \subset V(\Delta,\sigma)$ be the set of environments with change in mean $\Delta$ and Gaussian random noise with variance $\sigma^2$. Then, for $T \geq  \frac{\sigma^2}{\Delta^2}  \log \left(\frac{1}{2}(\lfloor1/2\eta\rfloor-1)\right)$, we have
    \begin{align*}
        \inf_{\pi \in \Pi} \sup_{v \in \Bar{V}} \mathbb{P}_{v,\pi} (|\hat{x}_T - x^*_v| \geq \eta) \geq& \\
        \frac{1}{8}\exp \bigg[-\frac{\Delta^2}{2\sigma^2}T + \frac{1}{2} &\log \left(\frac{1}{2}\left(\left\lfloor\frac{1}{2\eta}\right\rfloor-1\right)\right)\bigg].
    \end{align*}
\end{theorem}
\begin{proof}
    Fix some $\pi \in \Pi$. Let's then consider two cases.\\
    
    \textbf{First Case:} Suppose that 

    $$\forall v \in V, \quad \mathbb{P}_{v,\pi} (|\hat{x}_T - x^*_v| < \eta) \geq 1/2$$

    Then, from Theorem \ref{theorem_lb}, under this assumption, we have 

    $$\sup_{v \in \Bar{V}} \mathbb{P}_{v,\pi} (|\hat{x}_T - x^*_v| \geq \eta) \geq \frac{1}{8}\exp \left(-\frac{\Delta^2}{2\sigma^2}T + \frac{1}{2} \log \left(\frac{1}{2}(\lfloor1/2\eta\rfloor-1)\right)\right),$$

    as required.\\

    \textbf{Second Case:} Now, suppose instead that 

$$\exists v \in V, \quad \mathbb{P}_{v,\pi} (|\hat{x}_T - x^*_v| < \eta) < 1/2.$$

We can equivalently write that 

$$\exists v \in V, \quad \mathbb{P}_{v,\pi} (|\hat{x}_T - x^*_v| \geq \eta) \geq 1/2$$

And if this is the case, then we have the following
\begin{align*}
    \sup_{v \in \Bar{V}} \mathbb{P}_{v,\pi} (|\hat{x}_T - x^*_v| \geq \eta) &\geq \frac{1}{2}\\
    & \geq \frac{1}{8}\exp \left(-\frac{\Delta^2}{2\sigma^2}T + \frac{1}{2} \log \left(\frac{1}{2}(\lfloor1/2\eta\rfloor-1)\right)\right)
\end{align*}
As required. Where the final inequality holds since we are assuming that $T \geq \frac{\sigma^2}{\Delta^2}  \log \left(\frac{1}{2}(\lfloor1/2\eta\rfloor-1)\right)$.
    
\end{proof}

\newpage
\section*{Proofs For Small Budgets}
\section{Proofs For Small Budget Lower Bound} \label{app_smallT_lowerbound_proofs}
\subsection{Proof of Theorem \ref{lower_bound_exp_form}}
\begin{lemma}
 \label{covering lower bound fixed budget}
     Let $\Pi$ be the set of policies with fixed budget $T$. Let $\Bar{V} \subset V(\Delta,\sigma)$ be the set of environments with change in mean $\Delta$ and Gaussian random noise with variance $\sigma^2$. Denote $x^*_v$ as the change point in environment $v$. Then it holds that,
   $$\inf_{\pi \in \Pi} \sup_{v \in \Bar{V}} \, \mathbb{P}_{\pi,v}(|\hat{x}_T-x^*_v| > \eta) \geq 1- \frac{\Delta^2 T/2\sigma^2 + \log(2) }{\log(\lfloor\frac{1}{2\eta}\rfloor)}.$$   
\end{lemma}
\begin{proof}

This proof follows closely to that demonstrated in \cite{wainwright_2019}. Also, as it is oftentimes less obvious in this section, we will include all subscripts to denote the policy in use and environment in question. Additionally, we add superscript  for the random variable of the final estimate for the change point by policy $\pi$ in environment $v$ as $\hat{x}_T^{\pi,v}$.\\ 

Let $\eta \leq 0.5$ and  fix some arbitrary policy $\pi \in \Pi$. We choose $M$ environments $v_j \in \Bar{V}$ such that the $2\eta$ neighborhoods around the change point in each environment, $x^*_{v_j}$, are pairwise disjoint.  We can choose the covering number, $M$, to be at least $M = \lfloor1/2\eta\rfloor$. Let $J\sim \text{Uniform}\{1,...,M\}$. We can then show the following inequality, since we are taking a supremum on the left, which is greater than an average across any subset.

\begin{align}
    \sup_{v \in V} \, \mathbb{P}_{\pi,v}(|\hat{x}_T^{\pi,v}-x^*_v| > \eta) &\geq \frac{1}{M} \sum_{j=1}^{M} \mathbb{P}_{\pi,v_j}(|\hat{x}_T^{\pi,v_j}-x^*_{v_j}| > \eta)\nonumber\\
    & = \mathbb{P}_{J,\pi,v_J}(|\hat{x}_T^{\pi,v_J}-x^*_{v_J}| > \eta)\label{smallT_lb_proof_eqn1}\\
    \intertext{Where we denote $\mathbb{P}_{J,\pi,v_J}$ as the joint measure between $J\sim \text{Uniform}\{1,...,M\}$ and $\mathbb{P}_{\pi,v_J}$. Now,given our choice of $v_j$'s and by defining the test $\psi(\hat{x}_T^{\pi,v_J}) = \text{argmin}_{i \in \{1,...,M\}} |\hat{x}_T^{\pi,v_J} - x^*_{v_i}|$, we can then lower bound \eqref{smallT_lb_proof_eqn1} with}
    \sup_{v \in V} \, \mathbb{P}_{\pi,v}(|\hat{x}_T^{\pi,v}-x^*_v| > \eta) & \geq \mathbb{P}_{J,\pi,v_J}(\psi(\hat{x}_T^{\pi,v_J})\neq J)\nonumber\\
    & \geq 1- \frac{\frac{1}{M^2}\sum_{j,k=1}^M KL(\mathbb{P}_{\pi,v_j},\mathbb{P}_{\pi,v_k}) + \log(2)}{\log(M)} \label{smallT_lb_proof_eqn2}\\
    &\geq 1- \frac{\left(\frac{\Delta^2 T}{2 \sigma ^2}\right)+ \log(2)}{\log(M)} \label{smallT_lb_proof_eqn3}
\end{align}
Where equation \eqref{smallT_lb_proof_eqn2} holds from using  Fano's inequality and the loose bound for mutual information shown in equations 15.31 and 15.34 in \cite{wainwright_2019} respectively. Then equation \eqref{smallT_lb_proof_eqn3} holds from noting that, for any two environments in the sum in \eqref{smallT_lb_proof_eqn3}, the biggest difference in mean between the reward distributions anywhere in the action space is at most $\Delta$. Hence, using a similar idea to Lemma \ref{lemma_lb_cont_KL} and \ref{lemma_lb_cont_KL2}, we can use the divergence decomposition from \cite{BanditAlgosBook} to show that for any pair $i\neq k$ we have $KL(\mathbb{P}_{\pi,v_j},\mathbb{P}_{\pi,v_k}) \leq \frac{\Delta^2}{2\sigma^2}\mathbb{E}_{\pi,v_j}[T] = \frac{\Delta^2}{2\sigma^2}T$ in our fixed budget setting.

The proof is completed by noticing that for a fixed value of $\eta$, we can choose a covering of $[0,1]$ with covering number at least $M = \lfloor1/2\eta\rfloor$.\\
\end{proof}

We then simply extend Lemma \ref{covering lower bound fixed budget} by using the fact that $e^{-x} \leq 1-x/2$ when we have that $0 \leq x \leq 1.59$. Doing so gives us the required bound for Theorem \ref{lower_bound_exp_form}

\newpage
\section{Proofs For Small Budget Upper Bound}\label{app_smallT_upperbound_proofs}
\subsection{Proof for Theorem \ref{fixed budget upper bound}}

We can reuse a lot of the calculations we performed when analysing the backtracking algorithm, since we have defined the left and right elimination criteria in the same way. The different this time, is that we have to bound the probability that a good even occurs in every phase and we eliminate the correct half of the space. We define the good event in this case, event $G'_j$, as
\begin{equation*}
        G'_j = 
    \left\{  E_{L,j}^C \cap \left\{x^* \in [a_2^j,a_3^j) \right\}  \right\}  
    \cup \left\{   E_{R,j}^C \cap\left\{x^* \in  [a_1^j,a_2^j)\right\} \right\}.   
\end{equation*}

First, note that using the previous Appendix \ref{app_largeT_upper_proofs}, we can attain the following upper bound on related events.

\begin{lemma}\label{lemma_smallT_ub_proof} 
Given the rewards and actions from previous phases and under Algorithm \ref{alg:fixedbudget}; the probability that we fail to eliminate the correct half of the action space when it is actually on the right, and respectively left, can be upper bounded with
    \begin{align*}
        \mathbb{P}_j\left(E_{L,j} \bigg| x^* \in [a_2^j,a_3^j) \right) &\leq 2\exp \left(-\frac{t_j\Delta^2}{12\sigma^2}\right),\\
        \mathbb{P}_j\left(E_{R,j} \bigg| x^* \in [a_2^j,a_3^j) \right) &\leq 2\exp \left(-\frac{t_j\Delta^2}{12\sigma^2}\right).
    \end{align*}
\end{lemma}
\begin{proof}
    As mentioned before, for the first equation, we can directly use the calculation in equation \eqref{eqn_used_for_smallT_ub} and use an almost identical method for the second inequality. 
\end{proof}

Then, under Algorithm \ref{alg:fixedbudget}, we can upper bound the probability that event $G'_j$ fails to occur, given the actions and rewards from previous phases.

\begin{lemma}\label{lemma_smallT_ub_proof2}

Under Algorithm \ref{alg:fixedbudget}

$$\mathbb{P}_j(G_j') \geq 1-2\exp \left(-\frac{t_j\Delta^2}{12\sigma^2}\right).$$
\end{lemma}
\begin{proof}
    We can lower bound $\mathbb{P}_j(G_j')$ with
    \begin{align*}
        \mathbb{P}_j(G_j') &= \mathbb{P}_j\left(\left\{  E_{L,j}^C \cap \left\{x^* \in [a_2^j,a_3^j) \right\}  \right\}  
    \cup \left\{   E_{R,j}^C \cap\left\{x^* \in  [a_1^j,a_2^j)\right\} \right\}\right)\\
    &\geq \min \left(\mathbb{P}_j\left(  E_{L,j}^C \cap \left\{x^* \in [a_2^j,a_3^j) \right\}  \right), \mathbb{P}_j\left(   E_{R,j}^C \cap\left\{x^* \in  [a_1^j,a_2^j)\right\} \right)\right)\\
    &\geq \min \left(\mathbb{P}_j\left(  E_{L,j}^C \bigg|x^* \in [a_2^j,a_3^j)  \right), \mathbb{P}_j\left(   E_{R,j}^C \bigg|x^* \in  [a_1^j,a_2^j) \right)\right)\\
    &  \geq 1-2\exp \left(-\frac{t_j\Delta^2}{12\sigma^2}\right)
    \end{align*}
    As required. Where the final line comes from using Lemma \ref{lemma_smallT_ub_proof}.
\end{proof}

We know that, under the event that $G'_j$ occurs in every phase $j \in \{1,...,J\}$, our objective $|\hat{x}_T-x^*|<\eta$ is attained. Hence, we can take a union bound over all the phases, combined with Lemma \ref{lemma_smallT_ub_proof2}, we attain the required bound for Theorem \ref{fixed budget upper bound}.

\newpage
\section{Change Points And Intuition For Elimination Criteria}  \label{appendix_cps}

\subsection{Estimation in Offline Change Point Analysis}
Suppose we have a sequence of sample and observation pairs $\{x_i',y_i'\}_{i=1}^n$, ordered such that we have $x_1' \leq x_2' \leq ... \leq x_n'$. We also denote $\Bar{y}_{i:j}$ as the empirical mean of observations $y_i',...,y_j'$,
$$\Bar{y}_{i:j} = \sum_{k=i}^j y'_k.$$
Then, suppose we want to fit a piecewise constant model with two means $\mu_1$ and $\mu_2$ which change at some index $r$ in the sequence of observations $\{y_i'\}_{i=1}^n$, namely $f(x)=\mu_1\mathbb{I}\{x\leq x'_r\}+\mu_2\mathbb{I}\{x > x'_r\}$. To do so, we can write the residual sum of squares from fitting the model as
$$RSS(r) = \sum_{i=1}^{r}(y_i'-\hat{y}_1)^2 + \sum_{i=r+1}^{n}(y_i'-\hat{y}_2)^2.$$
From this, it is quick to check that the least squares estimators for the two means $\hat{y}_1$ and $\hat{y}_2$ are just $\Bar{y}_{1:r}$ and $\Bar{y}_{r+1:n}$ and we can therefore write the least squares estimator for the true change point index, to minimise this $RSS$, as
\begin{align}
    \hat{r} =\text{argmin}_{r \in \{1,..,n-1\}} \sum_{i=1}^{r}(y_i'-\Bar{y}_{1:r})^2 + \sum_{i=r+1}^{n}(y_i'-\Bar{y}_{r+1:n})^2 \label{change_points_LS_estimator}
\end{align}
It additionally turns out that this is equal to the Maximum Likelihood Estimator for the index of the change point in a sequence when we assume the underlying distributions of the observations are Gaussian \citep{ChenGuptaCP_Book}. Therefore we can use equation 2.7 in \citet{ChenGuptaCP_Book} to equivalently rewrite our least squares estimator from \eqref{change_points_LS_estimator} as  
\begin{align}
    \hat{r} =\text{argmax}_{r \in \{1,..,n-1\}} \frac{r(n-r)}{n} (\Bar{y}_{1:r}-\Bar{y}_{r+1:n})^2 \label{change_points_LS_estimator_new}
\end{align}

\subsection{Justification of Elimination Criteria}
In Algorithm \ref{alg:backtracking}, at the beginning of each phase we are given empirical means from repeatedly playing the five actions $0\leq a_1 <a_2<a_3\leq1$ (we drop the $j$ subscript for the actions in this section).

\textbf{The first problem} is to determine if a piecewise constant function fits better in terms of residual sum of squares when the change is between $a_1$ and $a_2$ or between $a_2$ and $a_3$. To do so, for simplicity, we choose to look only at the empirical means from playing actions $a_1,a_2,a_3$ (Note that we do not consider rewards from actions $0,1$ in this step only for simplicity. We could have instead included these actions on the boundary, but this would actually only affect the upper bound we attain in Theorem \ref{fixed budget upper bound backtracking} by at most a constant factor.). Hence by using equation \eqref{change_points_LS_estimator_new} and when fitting a piecewise constant model to our observed rewards, having the change point between $a_2$ and $a_3$ is a better fit that having the change point between $a_1$ and $a_2$ when we have 
\begin{equation}
    \frac{t_j(3t_j-t_j)}{3t_j} \left(\hat{\mu}_{a_1}-\frac{\hat{\mu}_{a_2}+\hat{\mu}_{a_3}}{2}\right)^2 < \frac{2t_j(3t_j-2t_j)}{3t_j} \left(\frac{\hat{\mu}_{a_1}+\hat{\mu}_{a_2}}{2} - \hat{\mu}_{a_3}\right)^2
\end{equation}
It turns out that this is equivalent to the event $E_{R,j}$, namely 

$$ \iff |\hat{\mu}_{a_1} - \hat{\mu}_{a_2}| < |\hat{\mu}_{a_2} - \hat{\mu}_{a_3}|, $$
which is the criteria we used in Algorithm \ref{alg:backtracking} to decide that the change point is in $x^* \in [a_2,a_3)$ and to then eliminate the region $[a_1,a_2)$. We can use a similar argument to justify the construction of the complementary event $E_{L,j}$.

Now, \textbf{the second problem} is to  determine if having the change \emph{outside} of the region $[a_1,a_3)$ would actually be a better fit. This time, again for simplicity, we consider only rewards from actions $0,a_1,a_3,1$ (where omitting $a_2$ helps make the backtracking condition simpler and the additional inclusion of $a_2$ would only improve the upper bound we attain in Theorem \ref{fixed budget upper bound backtracking} by at most a constant factor). In this case, comparing the regions $[0,a_1)$, $[a_1,a_3)$, $[a_3,1)$, we have that $[a_1,a_3)$ is \emph{not} the best fitting region for the change point when 

\begin{equation*}
    \left|\frac{\hat{\mu}_{0,t_j}+ \hat{\mu}_{a_1^j,t_j}}{2} - \frac{\hat{\mu}_{a_3^j,t_j}+\hat{\mu}_{1,t_j}}{2}\right| < \sqrt{\frac{3}{4}}\max \left( \left| \frac{\hat{\mu}_{0,t_j} + \hat{\mu}_{a_1^j,t_j} + \hat{\mu}_{a_3^j,t_j}}{3} - \hat{\mu}_{1,t_j}\right| ,\,  \left| \hat{\mu}_{0,t_j} - \frac{\hat{\mu}_{a_1^j,t_j} + \hat{\mu}_{a_3^j,t_j} + \hat{\mu}_{1,t_j}}{3}\right|     \right), 
\end{equation*}
which again comes from equation \eqref{change_points_LS_estimator_new}. We then define this as our criteria for backtracking in the definition of event $E_{P,j}$ except we modify the constant $\sqrt{\frac{3}{4}}$ to be $\frac{3}{4}$. We make this modification for simplicity as well as the fact that it allows us to construct a very slightly tighter upper bound in Theorem \ref{fixed budget upper bound backtracking}. This constant determines how strict we are with the backtracking procedure and it would be interesting to study exactly how this constant affects our upper bound for the failure probability. Perhaps in doing so we could find a way to optimise the strictness of our backtracking rule to further improve our upper bound. However this is beyong the scope of our work for now.

\newpage
\section{Proofs For Adaptive Algorithm} \label{Appendix_adaptive_algorithm}

\subsection{Proof of Theorem \ref{adaptive upper bound}}

Throughout this proof, we denote $C_1 = 1/600$, $C_2 = 13$.\\

Lets consider an environment $v$ with change in mean reward at the change point $\Delta$ and sub-Gaussian constant $\sigma^2$. We note that the event $\{T\geq\tau\}$, with
$$\tau = \gamma \frac{\sigma^2}{\hat{\Delta}^2_L} \log\left(\frac{1}{\eta}\right),$$
is equivalent to the event $\{\hat{\Delta}_L\geq\theta_\Delta\}$, where
\begin{equation} \label{eqn_adaptive_delta_threshold}
    \theta_\Delta = \sqrt{\gamma \frac{\sigma^2}{T} \log\left(\frac{1}{2\eta}\right)}.
\end{equation}
Now, we can use the Law of Total Probability 
to rewrite the failure probability from SHA as follows.

\begin{align}
    \mathbb{P}_{SHA,v} (|\hat{x}_T^{SHA} - x^*|> \eta) = &\mathbb{P}_{SHA,v} (|\hat{x}_T^{SHA} - x^*|> \eta | \hat{\Delta}_L \geq \theta_\Delta) \mathbb{P}_{SHA,v} (\hat{\Delta}_L \geq \theta_\Delta) \nonumber\\ 
    &+ \mathbb{P}_{SHA,v} (|\hat{x}_T^{SHA} - x^*|> \eta | \hat{\Delta}_L < \theta_\Delta)\mathbb{P}_{SHA,v} (\hat{\Delta}_L < \theta_\Delta)\nonumber\\
= & \mathbb{P}_{SHA,v} (|\hat{x}_{T-L}^{SHB} - x^*|> \eta| \hat{\Delta}_L \geq \theta_\Delta) \mathbb{P}_{SHA,v} (\hat{\Delta}_L \geq \theta_\Delta)\label{eqn_adaptive_conditions_implication}\\ 
    &+ \mathbb{P}_{SHA,v} (|\hat{x}_{T-L}^{SH} - x^*|> \eta| \hat{\Delta}_L <\theta_\Delta)\mathbb{P}_{SHA,v} (\hat{\Delta}_L < \theta_\Delta)\nonumber\\
    = & \mathbb{P}_{SHA,v} (|\hat{x}_{T-L}^{SHB} - x^*|> \eta) \mathbb{P}_{SHA,v} (\hat{\Delta}_L \geq \theta_\Delta)\label{eqn_adaptive_proof_failprob_decomposition}\\ 
    &+ \mathbb{P}_{SHA,v} (|\hat{x}_{T-L}^{SH} - x^*|> \eta)\mathbb{P}_{AHS,v} (\hat{\Delta}_L < \theta_\Delta)\nonumber
\end{align}

Where equation \eqref{eqn_adaptive_conditions_implication} holds since if the event $\{\hat{\Delta}_L\geq\theta_\Delta\}$ holds, then the event $\{T\geq\tau\}$ holds. Hence the SHA algorithm will run SHB for the final $T-L$ rounds and the estimate $\hat{x}_{T-L}^{SHB}$ will be returned (See Algorithm \ref{alg:adaptive} for explicit algorithm statement). Similarly, if event $\{\hat{\Delta}_L < \theta_\Delta\}$ holds, SHA will use SH for the final $T-L$ rounds. Equation \eqref{eqn_adaptive_proof_failprob_decomposition} comes from the fact that the random variable $\hat{\Delta}_L$ from the first $L$ rounds is independent to the estimate from SHB using the final $T-L$ rounds $\hat{x}_{T-L}^{SHB}$ and so the condition can be removed. Similarly, $\hat{\Delta}_L$ is independent to the estimate from SH $\hat{x}_{T-L}^{SH}$.

Hence from equation \eqref{eqn_adaptive_proof_failprob_decomposition}, the failure probability for SHA decomposes into the failure probability when using SHB for $T-L$ rounds multiplied by the probability that SHB is chosen, plus the failure probability from using SH multiplied by the probability that SH is chosen.

We now want to show that the failure probability for the SHA algorithm is near optimal for all budgets both small and large. We therefore split our analysis of the failure probability of SHA into three cases. \textbf{Case 1:} small budgets, \textbf{Case 2:} very large budgets, and \textbf{Case 3:} moderate budgets, which we define later. We demonstrate that for small budgets the probability that SHA chooses to run SH for the final $T-L$ rounds is high enough that the failure probability is of near optimal form. For very large budgets we show that SHA chooses to run SHB for the final $T-L$ rounds with high probability, and therefore our failure probability is near optimal. For moderate budgets we show that the failure probability from using SH or SHA would be similar and therefore our failure probability is near optimal regardless of our choice between the algorithms.

The result from Case 1 gives us the first upper bound of Theorem \ref{adaptive upper bound} for $T<T_1$ and combining the results from Case 2 and 3 gives us the second upper bound from Theorem \ref{adaptive upper bound} for $T\geq T_1$.\\

\textbf{Case 1: Small budget setting:} Suppose we are in the small budget setting where $T<T_1$, which can be equivalently written as

\begin{equation}\label{eqn_adaptive_case1_condition}
    \Delta < \sqrt{\frac{\sigma^2}{T}\left(1.59\log\left(\left\lfloor \frac{1}{2\eta}\right\rfloor\right) - 2\log(2)\right)}.
\end{equation}

We firstly note that by the assumption that the reward distributions for all of our actions are $\sigma^2$ sub-Gaussian, we know that the empirical means $\hat{\mu}_{1,L}$ and $\hat{\mu}_{0,L}$ are each $2\sigma^2/L$ sub-Gaussian. Hence, $\hat{\mu}_{1,L} - \hat{\mu}_{0,L}$ is $4\sigma^2/L$ sub-Gaussian.  We can then use Hoeffding's inequality (from Proposition 2.5 in \cite{wainwright_2019}) to show that, under the SHA algorithm our estimate for the size of the change in mean $\hat{\Delta}_L$ is concentrated as follows.

\begin{align}
    \mathbb{P}_{SHA,v}(\hat{\Delta}_L \geq \theta_\Delta) &= \mathbb{P}_{SHA}(|\hat{\mu}_{1,L} - \hat{\mu}_{0,L}| \geq \theta_\Delta) \nonumber\\
    &= \mathbb{P}_{SHA}(|\hat{\mu}_{1,L} - \hat{\mu}_{0,L}-\Delta+\Delta| \geq \theta_\Delta) \nonumber\\
    &\leq\mathbb{P}_{SHA}(|\hat{\mu}_{1,L} - \hat{\mu}_{0,L}-\Delta| +\Delta \geq \theta_\Delta) \nonumber\\
    &= \mathbb{P}_{SHA}(|\hat{\mu}_{1,L} - \hat{\mu}_{0,L}-\Delta| \geq \theta_\Delta - \Delta)\nonumber\\
    &\leq 2\exp\left(-\frac{L(\theta_\Delta-\Delta)^2}{8\sigma^2}\right)\label{eqn_adaptive_threshold_prob}
\end{align}

Note that by assumption $\gamma>1.59$, therefore
the $\theta_\Delta - \Delta$ term above is positive and this allows Hoeffding's inequality to hold for equation \eqref{eqn_adaptive_threshold_prob}. Hence, returning to the equation \eqref{eqn_adaptive_proof_failprob_decomposition} we have that the failure probability of SHA under small budgets is as follows. 

\begin{align}
    \mathbb{P}_{SHA,v} (|\hat{x}_T^{SHA} - x^*|> \eta) \leq & \mathbb{P}_{SHA,v} (|\hat{x}_{T-L}^{SHB} - x^*|> \eta) \mathbb{P}_{SHA,v} (\hat{\Delta}_L \geq \theta_\Delta)\nonumber\\ 
    &+ \mathbb{P}_{SHA,v} (|\hat{x}_{T-L}^{SH} - x^*|> \eta)\mathbb{P}_{AHS,v} (\hat{\Delta}_L < \theta_\Delta) \nonumber\\
    \leq & \mathbb{P}_{SHA,v} (|\hat{x}_{T-L}^{SHB} - x^*|> \eta) \mathbb{P}_{SHA,v} (\hat{\Delta}_L \geq \theta_\Delta)\nonumber\\ 
    &+ \mathbb{P}_{SHA,v} (|\hat{x}_{T-L}^{SH} - x^*|> \eta) \nonumber\\
    \leq& 2\exp\left(-C_1\frac{\Delta^2}{\sigma^2}(T-L) + C_2 \log(1/2\eta) -\frac{L(\theta_\Delta-\Delta)^2}{8\sigma^2}\right)\label{eqn_adaptive_subbing_probs}\\ 
    &+ 2\left\lceil \log _2 \left(\frac{1}{2\eta}\right) \right\rceil \exp\left(\frac{-(T-L)\Delta^2}{36\sigma^2 \log_2(1/2\eta)}\right) \nonumber\\
    \leq& 2\exp\left(-C_1\frac{\Delta^2}{\sigma^2}(T-L) + C_2 \log(1/2\eta) -B(\sqrt{\gamma}-\sqrt{1.59})^2\frac{\log\left( \frac{1}{2\eta}\right)}{8}\right)\label{adaptive_proof_eqn1}\\ 
    &+ 2\left\lceil \log _2 \left(\frac{1}{2\eta}\right) \right\rceil \exp\left(\frac{-(T-L)\Delta^2}{36\sigma^2 \log_2(1/2\eta)}\right) \nonumber\\
    \leq&2\exp\left(-C_1\frac{\Delta^2}{\sigma^2}(T-L)\right)\label{adaptive_proof_eqn2}\\ 
    &+ 2\left\lceil \log _2 \left(\frac{1}{2\eta}\right) \right\rceil \exp\left(\frac{-(T-L)\Delta^2}{36\sigma^2 \log_2(1/2\eta)}\right) \nonumber\\
    \leq& 4\left\lceil \log _2 \left(\frac{1}{2\eta}\right) \right\rceil \exp\left(\frac{-(T-L)\Delta^2}{\sigma^2} \min \left\{\frac{1}{36\log_2(1/2\eta)}, \, C_1\right\}\right) \nonumber\\
    \leq& 4\left\lceil \log _2 \left(\frac{1}{2\eta}\right) \right\rceil \exp\left(\frac{-(T-L)\Delta^2}{\sigma^2 \log_2(1/2\eta)} \min \left\{\frac{1}{36}, \, C_1\right\}\right) \nonumber\\
    \leq& 4\left\lceil \log _2 \left(\frac{1}{2\eta}\right) \right\rceil \exp\left(-C_1\frac{(T-L)\Delta^2}{\sigma^2 \log_2(1/2\eta)}\right) \label{equation_adaptive_case1_final_bound}
\end{align}

Note that equation \eqref{eqn_adaptive_subbing_probs} comes from substituting in our upper bounds from Theorem \ref{fixed budget upper bound backtracking}, Theorem \ref{fixed budget upper bound} and the concentration inequality from equation \eqref{eqn_adaptive_threshold_prob}. Also, in order to attain
\eqref{adaptive_proof_eqn1}, we consider lower bounding the quantity $(\theta_\Delta-\Delta)^2$. In Case 1 we have $\theta_\Delta>\Delta$ and an upper bound \eqref{eqn_adaptive_case1_condition} for $\Delta$. Hence, to upper bound equation \eqref{eqn_adaptive_subbing_probs} we substitute in the largest value $\Delta$ can take in Case 1, namely
$$\Delta = \sqrt{\frac{\sigma^2}{T}\left(1.59\log\left(\left\lfloor \frac{1}{2\eta}\right\rfloor\right) - 2\log(2)\right)},$$
as well as our definition of $\theta_\Delta$ from \eqref{eqn_adaptive_delta_threshold} and the fact that $L=BT$ in order to attain equation \eqref{adaptive_proof_eqn1}.
Then equation \eqref{adaptive_proof_eqn2} only holds when $B(\sqrt{\gamma}-\sqrt{1.59})^2/8>C_2$, which is assumed in the  statement of Theorem \ref{adaptive upper bound}. 
The upper bound in equation \eqref{equation_adaptive_case1_final_bound} is of the form of the small budget lower bound in Theorem \ref{lower_bound_exp_form}, hence SHA is nearly minimax optimal up to constants and loglog terms under small budgets.\\

\textbf{Case 2: Very Large Budgets:} Now assume that the budget is sufficiently large such that

\begin{equation*}
    \Delta > 2\theta_\Delta,
\end{equation*}
where $\theta_\Delta$ is defined in equation \eqref{eqn_adaptive_delta_threshold}. Note that in this case, we can make a similar concentration argument to Case 1, using Hoeffding's inequality to show that $\hat{\Delta}_L < \theta_\Delta$ occurs with small probability. In particular,

\begin{align}
    \mathbb{P}_{SHA,v}(\hat{\Delta}_L < \theta_\Delta) &= \mathbb{P}_{SHA}(|\hat{\mu}_{1,L} - \hat{\mu}_{0,L}| <\theta_\Delta) \nonumber\\
    &= \mathbb{P}_{SHA}(|\hat{\mu}_{1,L} - \hat{\mu}_{0,L}-\Delta+\Delta| < \theta_\Delta) \nonumber\\
    &\leq\mathbb{P}_{SHA}(|\Delta| - |\hat{\mu}_{1,L} - \hat{\mu}_{0,L}-\Delta| < \theta_\Delta) \nonumber\\
    &= \mathbb{P}_{SHA}(|\hat{\mu}_{1,L} - \hat{\mu}_{0,L}-\Delta| > \Delta - \theta_\Delta)\nonumber\\
    &\leq 2\exp\left(-\frac{L(\theta_\Delta-\Delta)^2}{8\sigma^2}\right).\label{eqn_adaptive_threshold_prob2}
\end{align}

Note that, since in Case 2 we assume that $\Delta > 2\theta_\Delta$, and therefore $\Delta > \theta_\Delta$. Hence the $\Delta - \theta_\Delta$ term above is positive and this allows Hoeffding's inequality to hold for equation \eqref{eqn_adaptive_threshold_prob2}.
We can now rewrite the failure probability for the SHA algorithm in this case as 

\begin{align}
    \mathbb{P}_{SHA,v} (|\hat{x}_T^{SHA} - x^*|> \eta) \leq & \mathbb{P}_{SHA,v} (|\hat{x}_{T-L}^{SHB} - x^*|> \eta) \mathbb{P}_{SHA,v} (\hat{\Delta}_L \geq \theta_\Delta)\nonumber\\ 
    &+ \mathbb{P}_{SHA,v} (|\hat{x}_{T-L}^{SH} - x^*|> \eta)\mathbb{P}_{AHS,v} (\hat{\Delta}_L < \theta_\Delta) \nonumber\\
    \leq & \mathbb{P}_{SHA,v} (|\hat{x}_{T-L}^{SHB} - x^*|> \eta) \nonumber\\ 
    &+ \mathbb{P}_{SHA,v} (|\hat{x}_{T-L}^{SH} - x^*|> \eta)\mathbb{P}_{AHS,v} (\hat{\Delta}_L < \theta_\Delta) \nonumber\\
    \leq& \exp\left(-C_1\frac{\Delta^2}{\sigma^2}(T-L) + C_2 \log(1/2\eta) \right)\label{eqn_adaptive_subbing_theorems2}\\ 
    &+ 4\left\lceil \log _2 \left(\frac{1}{2\eta}\right) \right\rceil \exp\left(\frac{-(T-L)\Delta^2}{36\sigma^2 \log_2(1/2\eta)} -\frac{L(\theta_\Delta-\Delta)^2}{8\sigma^2}\right) \nonumber\\
    \leq& \exp\left(-C_1\frac{\Delta^2}{\sigma^2}(T-L) + C_2 \log(1/2\eta) \right)\nonumber\\ 
    &+ 4\left\lceil \log _2 \left(\frac{1}{2\eta}\right) \right\rceil \exp\left(\frac{-(T-L)\Delta^2}{36\sigma^2 \log_2(1/2\eta)} -\frac{L(\Delta/2)^2}{8\sigma^2}\right) \label{adaptive_proof_eqn3}\\
    =& \exp\left(-C_1\frac{\Delta^2}{\sigma^2}(T-L) + C_2 \log(1/2\eta) \right)\nonumber\\ 
    &+ 4\left\lceil \log _2 \left(\frac{1}{2\eta}\right) \right\rceil \exp\left(\frac{-(T-L)\Delta^2}{36\sigma^2 \log_2(1/2\eta)} -\frac{L\Delta^2}{32\sigma^2}\right) \nonumber\\
    \leq& \exp\left(-C_1\frac{\Delta^2}{\sigma^2}(T-L) + C_2 \log(1/2\eta) \right)\nonumber\\ 
    &+ 4\left\lceil \log _2 \left(\frac{1}{2\eta}\right) \right\rceil \exp\left( -\frac{L\Delta^2}{32\sigma^2}\right) \nonumber\\
    \leq& 5\exp\left(-\frac{\Delta^2}{\sigma^2} \min\left\{(T-L)C_1,L/32\right\}+C_2\log(1/2\eta)\right) \nonumber\\
    \leq& 5\exp\left(-C_1 \frac{\Delta^2}{\sigma^2} B(1-B)T+C_2\log(1/2\eta)\right). \label{eqn_adaptive_case2_final_bound}
\end{align}

 Here equation \eqref{eqn_adaptive_subbing_theorems2} comes from substituting our upper bounds for SH in Theorem \ref{fixed budget upper bound} and SHB in Theorem \ref{fixed budget upper bound backtracking}, as well as using our upper bound in equation \eqref{eqn_adaptive_threshold_prob2}. Furthermore, equation \eqref{adaptive_proof_eqn3} comes from the fact that $\Delta > 2\theta_\Delta$. Finally, equation \eqref{eqn_adaptive_case2_final_bound} comes from recalling that $L=BT$.
 Note that equation \eqref{eqn_adaptive_case2_final_bound} matches our lower bound from Theorem \ref{corr_large_budget_lb} up to constants, hence SHA is near optimal up to constants for these very large budgets.

\textbf{Case 3 Moderate Budgets:} Let's suppose instead that the budget is only moderately large and is bounded by

\begin{equation*}
    \Delta \in \left[\sqrt{\frac{\sigma^2}{T}\left(1.59\log\left(\left\lfloor \frac{1}{2\eta}\right\rfloor\right) - 2\log(2)\right)},2\theta_\Delta\right].
\end{equation*}
We can equivalently write this condition as
\begin{equation}\label{eqn_adaptive_case3_condition}
    \Delta \in \left[\sqrt{\frac{\sigma^2}{T}\left(1.59\log\left(\left\lfloor \frac{1}{2\eta}\right\rfloor\right) - 2\log(2)\right)},2\sqrt{\gamma \frac{\sigma^2}{T} \log\left(\frac{1}{2\eta}\right)}\right].
\end{equation}

In this case, as long as we have $\gamma<1800/(1-B)$, then for all such $\Delta$ in the moderate budget setting \eqref{eqn_adaptive_case3_condition}, we have a better guarantee for SH in Theorem \ref{fixed budget upper bound} than for SHB in Theorem \ref{fixed budget upper bound backtracking} with budget input $T-L$. This is because, combining $\gamma<1800/(1-B)$ and our condition \eqref{eqn_adaptive_case3_condition} for Case 3, implies that $T-L<7200 \frac{\sigma^2}{\Delta^2} \log\left(\frac{1}{\eta}\right)$. Plugging in these values for the budget $T-L$ into Theorems \ref{fixed budget upper bound} and \ref{fixed budget upper bound backtracking} shows that our gaurantee for SH outperforms SHB in this moderate budget setting. We note that $7200 \frac{\sigma^2}{\Delta^2} \log\left(\frac{1}{\eta}\right)$ is different to the definition of $T_1$ (see Section \ref{section small budget}). The constant $7200$ comes from the comparison of our two upper bounds for SH/SHB (which are loose in constants), whereas in practice the budget at which SHB begins to outperform SH is significantly smaller than this (see Section \ref{appendix:more sims}).

Hence,  we can upper bound our failure probability in this region as

\begin{align}
    \mathbb{P}_{SHA,v} (|\hat{x}_T^{SHA} - x^*|> \eta) \leq & \mathbb{P}_{SHA,v} (|\hat{x}_{T-L}^{SHB} - x^*|> \eta) \mathbb{P}_{SHA,v} (\hat{\Delta}_L \geq \theta_\Delta)\nonumber\\ 
    &+ \mathbb{P}_{SHA,v} (|\hat{x}_{T-L}^{SH} - x^*|> \eta)\mathbb{P}_{AHS,v} (\hat{\Delta}_L < \theta_\Delta) \nonumber\\
    \leq & \mathbb{P}_{SHA,v} (|\hat{x}_{T-L}^{SHB} - x^*|> \eta) \cdot 1 \nonumber\\ 
    &+ \mathbb{P}_{SHA,v} (|\hat{x}_{T-L}^{SH} - x^*|> \eta) \cdot 1  \nonumber\\
    \leq& 2\exp\left(-C_1\frac{\Delta^2}{\sigma^2}(T-L) + C_2 \log(1/2\eta) \right)\label{eqn_adaptive_SH_less_SHB}\\
    =& 2\exp\left(-C_1\frac{\Delta^2}{\sigma^2}(1-B)T + C_2 \log(1/2\eta) \right).\label{eqn_adaptive_case3_final}
\end{align}

Here equation \eqref{eqn_adaptive_SH_less_SHB} comes from the above discussion that for moderate budgets in Case 3, our gaurantee for SH is better than that of SHB. Equation \eqref{eqn_adaptive_case3_final} comes from recalling that $L=BT$.
Equation \eqref{eqn_adaptive_case3_final} is of the form of our lower bound in Theorem \ref{corr_large_budget_lb}, therefore SHA algorithm is nearly optimal under moderate budgets as well.

\qed

\end{document}